\documentclass[preprint,12pt, sort&compress]{elsarticle}

\usepackage{amssymb}

\usepackage{amsmath}

\usepackage{subfig}

\usepackage{booktabs}

\usepackage{textcomp}

\usepackage[vlined,ruled]{algorithm2e}
\SetAlgoSkip{bigskip}

\journal{Robotics and Autonomous Systems}

\begin{document}

\begin{frontmatter}



\title{Methods for Stochastic Collection and Replenishment (SCAR) optimisation for persistent autonomy}


\author{Andrew W. Palmer, Andrew J. Hill, Steven J. Scheding \corref{cor1}}
\cortext[cor1]{Email address: \{a.palmer;a.hill;s.scheding\}@acfr.usyd.edu.au}

\address{Australian Centre for Field Robotics, The University of Sydney, NSW, Australia}

\begin{abstract}

Consideration of resources such as fuel, battery charge, and storage space, is a crucial requirement for the successful persistent operation of autonomous systems. The Stochastic Collection and Replenishment (SCAR) scenario is motivated by mining and agricultural scenarios where a dedicated replenishment agent transports a resource between a centralised replenishment point to agents using the resource in the field. The agents in the field typically operate within fixed areas (for example, benches in mining applications, and fields or orchards in agricultural scenarios), and the motion of the replenishment agent may be restricted by a road network. Existing research has typically approached the problem of scheduling the actions of the dedicated replenishment agent from a short-term and deterministic angle. This paper introduces a method of incorporating uncertainty in the schedule optimisation through a novel prediction framework, and a branch and bound optimisation method which uses the prediction framework to minimise the downtime of the agents. The prediction framework makes use of several Gaussian approximations to quickly calculate the risk-weighted cost of a schedule. The anytime nature of the branch and bound method is exploited within an MPC-like framework to outperform existing optimisation methods while providing reasonable calculation times in large scenarios.

\end{abstract}


\end{frontmatter}


\section{Introduction}

Achieving persistent autonomy requires careful management of finite resources such as fuel, battery charge, and storage space. While there are some scenarios where agents are able to continue operating by collecting energy from the environment, such as gliding Unmanned Aerial Vehicles (UAVs) exploiting thermals to remain airborne \cite{Nguyen2013}, in the majority of cases the agent uses a resource which must be replenished to enable persistence. Early work on replenishment involved the robot returning to a charging dock to facilitate autonomous recharging \cite{Austin2001, Luo2005}, which now forms the basis of recharging in some commercial systems such as the iRobot Roomba \cite{Forlizzi2006} and InTouch Health RP-7i \cite{RP-7i}. However, there are many cases in which the use of a dedicated replenishment agent can provide direct benefit. Examples include refuelling or recharging of UAVs in flight, refuelling of satellites in orbit, and data collection from underwater wireless sensor networks. Using a dedicated agent to replenish the resource of each agent in the field enables the agents to be smaller, simpler, and cheaper \cite{Zebrowski2005}. 

The Stochastic Collection and Replenishment (SCAR) problem was first introduced by the authors in \cite{Palmer2013} as a generic resource management scenario that is motivated by scenarios commonly found in mining and agricultural environments. Possible resources include fuel, battery charge, food, and water for the replenishment case, and electronic data, mined ore, and harvested fruit for the collection case. In these scenarios, a dedicated replenishment agent transports the resource between a centralised replenishment point and the agents operating in the field. The motion of the replenishment agent is typically restricted by roads or other obstacles, and the agents in the field operate within fixed areas such as benches in mining scenarios, and fields and orchards in agricultural scenarios. The SCAR scenario differs from similar research in several key ways. Firstly, the examined literature assumes that each agent is visited only once. When considering persistent scenarios, neglecting future visits may lead to sub-optimal decision making. Secondly, the replenishment agent is generally assumed to have sufficient capacity such that it will not run out of the resource. Again, this is not the case for persistent autonomy. Finally and most importantly, the examined literature assumes that the problem is deterministic. In practice, this is not the case as agent parameters such as speed and resource usage rate will have elements of uncertainty. 

A prediction framework for SCAR scenarios was developed by the authors in \cite{Palmer2013} which was able to quickly calculate the risk-weighted cost of a schedule for the replenishment agent. This was then used within an A* optimisation method in \cite{Palmer2014a} to schedule the actions of the replenishment agent in small SCAR scenarios. This paper introduces an improved version of the prediction framework and develops a branch and bound optimisation method for use in large SCAR scenarios where optimisation time is limited. The specific contributions of this paper include:

\begin{itemize}
	\item a novel framework for predicting the risk-weighted cost of a schedule;
	\item a novel Gaussian approximation to the inverse of a Gaussian distributed random variables;
	\item a novel Gaussian approximation to the generalised rectified Gaussian distribution;
	\item an anytime branch and bound optimisation method; and
	\item a computational comparison of the proposed approaches with the state of the art. 
\end{itemize}

The prediction framework is accurate and computationally efficient. This is significant as complex objective functions that incorporate uncertainty have, thus far, required the use of Monte Carlo simulation to calculate an accurate estimate of the expected value. Combined with the branch and bound approach, the proposed methods are shown to outperform the existing approaches in terms of minimising the downtime of the agents in the field.

The rest of this paper is structured as follows: Section \ref{s:rellit} discusses the related literature, and a description of the SCAR scenario is introduced in Section \ref{s:probdef}. The improved prediction framework is outlined in Section \ref{s:pred}, and approximations for using Gaussian distributed random variables within the prediction framework are developed in Section \ref{s:gaussian}. The optimisation methods are introduced in Section \ref{s:optmeth}. The prediction and optimisation methods are then compared in a computational study in Section \ref{s:compstud}, and conclusions and future work are presented in Section \ref{s:conc}. 

\section{Related literature} \label{s:rellit}

The main replenishment and collection scenarios in the literature are refuelling or aircraft and satellites, recharging of ground and aerial robots, and collecting data from wireless sensor networks. In general, these are framed as NP-hard combinatorial optimisation problems that resemble classical problems such as the restricted Travelling Salesman Problem (TSP) with time windows, or the parallel machine manufacturing job shop problem. 

The aerial refuelling problem was examined in \cite{Jin2006a, Kaplan2012, Kaplan2013, Barnes2004}. Jin et al. \cite{Jin2006a} used a recursive dynamic programming approach to determine the optimal order for refuelling the aircraft to minimise the priority weighted time of refuelling for each aircraft. Such an approach is not applicable to SCAR scenarios as the optimal future decision in a SCAR scenario is dependent on the sequence of decisions leading to it. Kaplan and Rabadi \cite{Kaplan2012, Kaplan2013} developed heuristic and meta-heuristic approaches. In \cite{Kaplan2012} they found that the meta-heuristic was outperformed by the heuristic in large scenarios, and in \cite{Kaplan2013} they used the heuristic to generate an initial schedule for the meta-heuristic to improve performance. 

An assumption of the work of Jin et al. and Kaplan and Rabadi was that the aircraft being refuelled were in formation behind the tanker aircraft, meaning that the time between refuelling each aircraft was determined purely by the set-up time required by that particular aircraft. Barnes et al. \cite{Barnes2004} considered the inter-theatre aerial refuelling, where the flight time of the tanker aircraft between aircraft to be refuelled was treated as sequence-dependent set-up times. This is similar to how the travel between satellites was treated for refuelling a constellation of satellites \cite{Shen2002}. The solution methods used in these cases were heavily tailored towards the specific scenario under consideration and are not generalisable to SCAR scenarios. 

Recharging of ground and aerial robots was examined in \cite{Kannan2013, Litus2009, Mathew2015}. Kannan et al. \cite{Kannan2013} developed a market-based solution for determining a good recharging strategy, but only examined scenarios with a single user agent, while Litus et al. \cite{Litus2009} focussed on determining rendezvous locations given an optimal recharging order. Mathew et al. \cite{Mathew2015} developed a receding horizon approach for simultaneously calculating a recharging order and rendezvous locations. These papers assumed that the recharging agent had sufficient capacity to fully recharge all of the user agents without itself having to return to a charging point, ignoring a critical aspect for persistent operation. 

The literature on collection scenarios has focused on the use of data mules to collect data from wireless sensor networks. This strategy can be particularly useful for underwater wireless sensor networks, used for long-term monitoring of coral reefs, where wireless communications over long distances can be difficult. Vasilescu et al. \cite{Vasilescu2005} used an Autonomous Underwater Vehicle (AUV) to travel to each sensor node and retrieve the collected data. Similar scenarios are presented in \cite{Tirta2004, Yuan2007, Tekdas2009, Tekdas2012, Bhadauria2011, Ma2013, Yan2014}. The problem is generally treated as a variant of the TSP with the aim of minimising the total distance travelled, the data latency, or the total schedule time. 

Most of the above literature does not approach this problem from a persistent autonomy perspective---the optimisation methods generally consider visiting each agent or node only once, assume that the replenishment agent has sufficient or unlimited capacity, and ignore time-varying effects such as variable replenishment times and deadlines. In addition, all of the above literature bar \cite{Mathew2015} ignore uncertainty. In \cite{Mathew2015}, arbitrary safety margins were used to reduce plan failure, with no consideration of the actual risk associated with each task. 

Uncertainty in scheduling has received limited study, even in the classical manufacturing scheduling literature, due to its difficulty in comparison to deterministic problems \cite{Allahverdi2008}. Typical approaches to incorporating uncertainty include chance constraints \cite{Fang2014a}, Monte Carlo simulation \cite{Raboin2014}, using conservative estimates of the uncertain parameters \cite{Bertsimas2003}, replanning frequently \cite{Yoon2007}, or simply ignoring the uncertainty \cite{Arnaout2006}. Of these approaches, only chance constraints and Monte Carlo simulation allow risk to be incorporated into the optimisation. 

The prediction framework presented in the authors' previous work \cite{Palmer2013} produced a similar result to Monte Carlo simulation. Instead of sampling the probability distributions, however, it used analytical and approximation methods to propagate the entire probability distribution, resulting in a calculation time that was orders of magnitude faster than the Monte Carlo approach. This was used in \cite{Palmer2014a} within an A* optimisation approach to evaluate each schedule under consideration. Using the prediction framework to incorporate risk was shown to outperform an A* approach that ignored the risk. While it was sufficient for small SCAR scenarios, this approach is infeasible for larger scenarios due to the size of the search space. 

Optimisation methods which are generally used in larger scenarios include heuristics, meta-heuristics, and anytime combinatorial optimisation methods. The Apparent Tardiness Cost (ATC) heuristic and simulated annealing meta-heuristic were used by Kaplan and Rabadi \cite{Kaplan2012} for the aerial refuelling problem. They found that the ATC heuristic outperformed simulated annealing in larger scenarios, and in later work combined the two methods to find better solutions \cite{Kaplan2013}. Anytime combinatorial optimisation methods include branch and bound, and anytime A* methods such as ARA*. ARA* uses inadmissible heuristics to quickly find sub-optimal solutions, before refining the solution by moving towards an admissible heuristic \cite{Ferguson2005}, while branch and bound prunes parts of the solution tree which have an estimated lower cost bound that is higher than the cost of the current solution \cite{Land1960}. Due to the difficulty in deriving heuristics that can be used in ARA*, branch and bound has been chosen as the solution method in this paper. 

\section{The SCAR scenario} \label{s:probdef}

\begin{figure}
	\centering
	\includegraphics[width=0.65\textwidth]{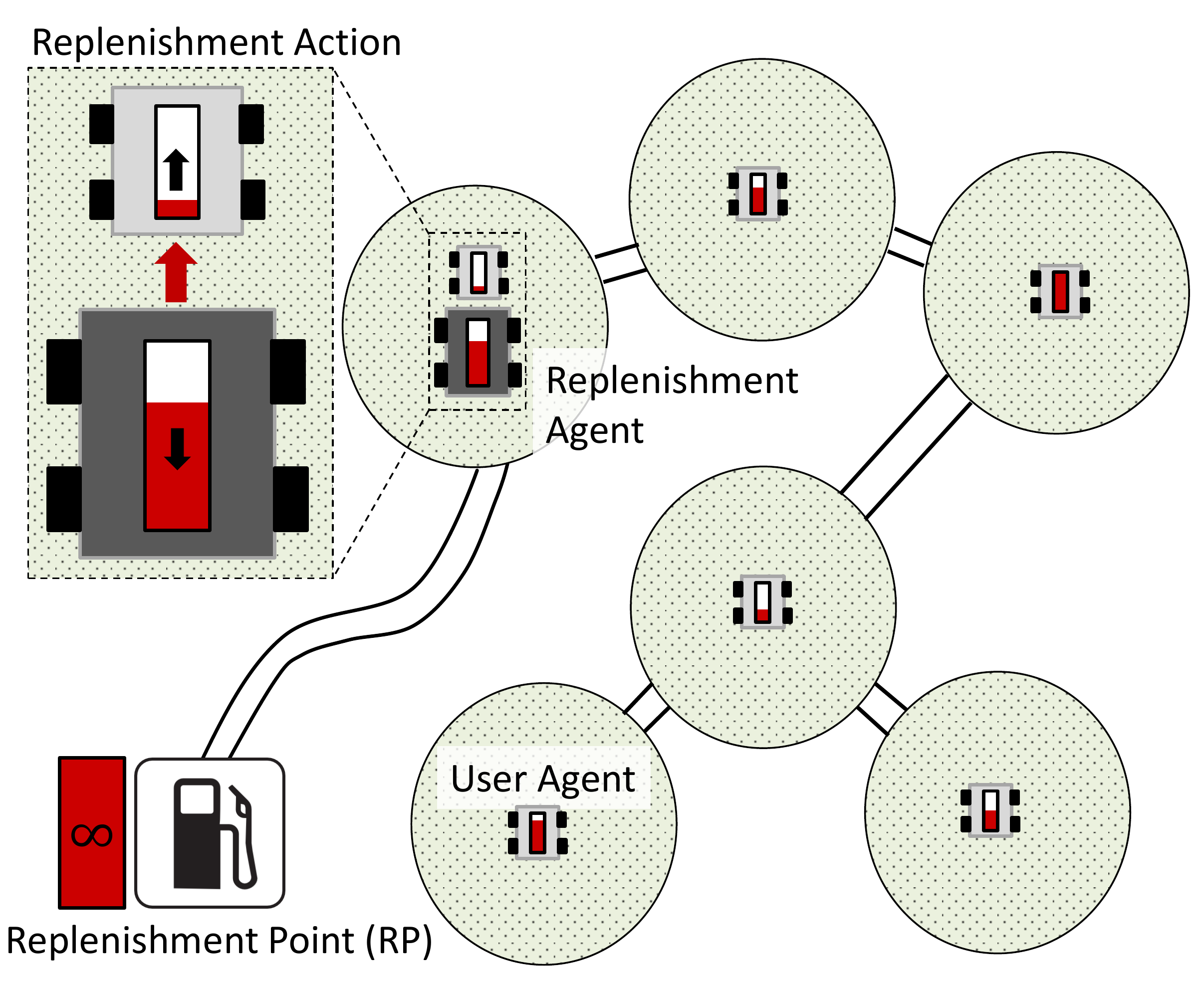}
	\caption{An example SCAR scenario. A replenishment agent (dark grey) travels to, and replenishes, the user agents (light grey) operating in the field. The replenishment agent must return to the replenishment point (bottom left) to replenish its supply of the resource. The replenishment point has infinite capacity of the resource. In this example, the travel of the replenishment agent is restricted by roads between the operational areas of the user agents. Inset: The replenishment agent transfers the resource to the user agent, diminishing its supply of the resource and increasing the resource reserves of the user agent. }
	\label{f:overview}
\end{figure}

SCAR scenarios are motivated by mining and agricultural scenarios such as replenishing excavators and drills using fuel and water trucks, and collecting harvested crops and picked fruit. These scenarios consist of multiple user agents, such as excavators and harvesters, that either consume or collect a resource over time. As they have a limited capacity of the resource, a dedicated replenishment agent, such as a fuel truck, rendezvous with the user agents and replenishes their supply of the resource. This enables the user agents to continue operating in the field without having to return to fixed replenishment infrastructure. Collection scenarios are identical to replenishment scenarios if the resource under consideration is storage space. An example SCAR scenario is shown in Figure \ref{f:overview}. 

The fleet of user agents is heterogeneous, and, for the scenarios under consideration, there is a centralised replenishment point such as a refuelling station or silo. The parameters of the user agents, replenishment agent, and replenishment point, such as their speed, set-up and pack-up times, and resource usage rates, are stochastic. The user agents operate in defined areas such as benches on a mine site for drills and excavators, and fields for tractors and harvesters in agricultural scenarios. The distances between the operational areas of the user agents are generally much larger than the size of those areas, and any variations in the travel times of the replenishment agents due to the movements of the user agents are assumed to be accounted by the uncertain travel speed, and set-up and pack-up times of the replenishment agent.

Note that user agents are not required to remain in their operational area indefinitely. User agents that move to new benches or fields within the scope of a schedule can be incorporated by treating the time that they leave their current location as the latest time that it can be replenished at that location. Similarly, the time that it arrives at its new location can be treated as the ready time for that agent. Given uncertainty in the arrival time of the replenishment agent, a probabilistic check can be used to evaluate whether the replenishment agent will arrive at the location before the deadline or after the ready time with sufficient probability for it to be allowed as a valid task. In this way, operational areas can be added or removed from the network. Locations along roads are not considered valid locations for the replenishment agent to service a user agent as blocking the road creates a hazardous situation. 

It is assumed that the user agents can recover from exhausting their supply of the resource, i.e. they enter a safe zero-resource state. This is modelled in this paper as a soft deadline which has a cost that increases linearly with the time that the user agents are not operational. A hard deadline would correspond to, for example, a UAV exhausting its supply of fuel mid-flight, causing the loss of the agent. 

The following subsections introduce the system parameters, variables, and constraints, and the optimisation objectives. Parameters and variables are treated as either known to a high degree of certainty (denoted by a lower case letter), or as random variables (denoted by an upper case letter). The uncertain parameters and variables in this paper are assumed to be Gaussian distributed random variables, and several approximations are introduced in Section \ref{s:gaussian} to facilitate the mathematical operations performed on them. Other probability distributions can be used provided that the appropriate operations and approximations exist. Units do not matter so long as they are consistent. 

\subsection{Parameters, variables, and constraints}

For each user agent, $i$, in an $n$-user agent system:

\begin{itemize}
	\item $ c_{u,i} $ is the resource capacity
	\item $ l_{u,i} $ is the current resource level, where $ l_{u,i} \in [0,c_{u,i}] $
	\item $ L_{u,i} $ is the estimated future resource level, where $ L_{u,i} \in [0,c_{u,i}] $
	\item $ R_{u,i} $ is the resource usage rate
	\item $ w_{i} $ is a user-defined weight or priority for the user agent
\end{itemize}

It is assumed that the user agents continue consuming the resource while being replenished. If the current resource level of the user agent reaches 0, then the user agent ceases operation and incurs downtime. 

For the replenishment agent:

\begin{itemize}
	\item $ c_{a} $ is the resource capacity
	\item $ l_{a} $ is the current resource level, where $ l_{a} \in [0,c_{a}] $
	\item $ L_{a} $ is the estimated future resource level, where $ L_{a} \in [0,c_{a}] $
	\item $ R_{a} $ is the resource replenishment rate into the user agent
	\item $ D_{sa} $ is the duration of time required for the replenishment agent to set-up at a user agent
	\item $ D_{pa} $ is the duration of time required for the replenishment agent to pack-up at a user agent
	\item $ V_{a} $ is the velocity
\end{itemize}

The replenishment agent is assumed to have a separate supply of fuel or battery charge for its own operation that is replenished in parallel at the replenishment point. It is also assumed to be able to service only one user agent at a time. To replenish a user agent, the replenishment agent must first travel to the user agent and set up before commencing the replenishment. After it has either fully replenished the user agent or exhausted its own supply of the resource, the replenishment agent must pack up before travelling to the next task. 

Finally, for the replenishment point:

\begin{itemize}
	\item $ R_{r} $ is the resource replenishment rate into the replenishment agent
	\item $ D_{sr} $ is the duration of time required for the replenishment agent to set-up at the replenishment point
	\item $ D_{pr} $ is the duration of time required for the replenishment agent to pack-up at the replenishment point
\end{itemize}

For the replenishment agent to be replenished, it must first travel to the replenishment point, then set up, be fully replenished by the replenishment point, and then pack up before moving onto the next task. It cannot service a user agent while being replenished by the replenishment point. Note that the set-up and pack-up times at the replenishment point are different to those at the user agents. 

The distance between the replenishment agent and user agent $i$ is denoted $ s_{au,i} $, and the distance between the replenishment agent and the replenishment point is denoted $ s_{ar} $. In the case where there are multiple routes between tasks, the replenishment agent is assumed to take the shortest (fastest) route.

\subsection{Optimisation}

The aim of the optimiser is to minimise the total downtime of they user agents by optimising the actions of the replenishment agent, where downtime is incurred when a user agent has exhausted its supply of the resource. More generally, the objective function is the total weighted downtime of the user agents, $\zeta$:

\begin{equation}\label{eq:objfunc}
\min \zeta = \sum\limits_{i = 1}^{n}w_{i}d_{c, i}
\end{equation}

where $d_{c,i}$ is the total downtime incurred by user agent $i$. Given the uncertainty present in the agent parameters, the total downtime that is expected to be incurred for a given schedule can only be estimated. Two methods for estimating the downtime of the user agents for a given schedule of actions are given in the authors' previous work \cite{Palmer2013}, and an improved method is developed in Section \ref{s:pred}. 

The total weighted downtime objective is only appropriate for planning over an infinite horizon. As has been demonstrated in the authors' previous work \cite{Palmer2014a}, the use of a ratio based objective function can produce better results when using combinatorial optimisation methods to optimise the actions of the replenishment agent. The ratio objective function, $\lambda$, is defined as:

\begin{equation}\label{eq:objfuncratio}
\min \lambda = \frac{\zeta}{n d_{m}}
\end{equation}
where $d_{m}$ is the total time for the schedule to be executed. Provided the weights sum to $n$, then the objective function is bound between 0 and 1:

\begin{equation}
\textrm{if } \sum_{i=0}^{n-1} w_{i} = n \textrm{ then } 0 \le \lambda \le 1
\end{equation}

The ratio objective function enables comparison between schedules that have the same number of tasks but take different lengths of time to execute. A schedule consists of an ordered list of tasks for the replenishment agent to perform. The task of visiting the replenishment point is denoted by 0, while the task of replenishing a user agent is denoted by its index number. An example schedule, $\boldsymbol{\theta}$, of a 4-user agent SCAR scenario is:

\[\boldsymbol{\theta} = (1,0,4,2,1,4)\]

In this schedule, the first task, $\boldsymbol{\theta}_{1}$, involves the replenishment agent replenishing the first user agent before visiting the replenishment point for task $\boldsymbol{\theta}_{2}$. For tasks $\boldsymbol{\theta}_{3}, \boldsymbol{\theta}_{4}, \boldsymbol{\theta}_{5} \textrm{ and } \boldsymbol{\theta}_{6}$, it would replenish agents 4, 2, 1 and 4 again in that order. Note that user agents 1 and 4 appear multiple times, and user agent 3 does not appear at all. Unlike existing replenishment scenarios, the assumptions of the SCAR scenario allow for user agents to be visited multiple times, or not at all within a given schedule. 

The optimisation is performed within a framework similar to Model Predictive Control (MPC)---the optimiser returns a new task or schedule after each task is completed. Thus, unexpected changes to the system state are incorporated into the optimisation each time a task is performed. 

\section{Prediction framework} \label{s:pred}

This section develops an improved version of the analytical prediction framework presented in \cite{Palmer2013}. The framework takes a schedule for the replenishment agent as input, and its aim is to predict the future resource levels of the user and replenishment agents, and to estimate the total weighted downtime of the user agents and the total time taken to execute the schedule. This framework will be used within the combinatorial optimisation methods developed in Section \ref{s:optmeth} to evaluate the objective function. 

In \cite{Palmer2013}, two continuous-time frameworks were developed---the first used a Monte Carlo approach to estimate the downtime of the user agents given the uncertainty in the system parameters, and the second used analytical and approximation methods to propagate the uncertainty instead of the sampling used in the Monte Carlo approach. The framework presented in Algorithm \ref{a:single} builds on the second approach from \cite{Palmer2013}. It is agnostic to the type of probability distribution used, and Section \ref{s:gaussian} presents approximations that enable the use of Gaussian distributed random variables. To use other probability distributions with this framework, these approximations would need to be formulated for the cases where the appropriate operations do not exist. 

\begin{algorithm}
	\DontPrintSemicolon
	\SetAlgoNoEnd
	\SetKwFunction{Analytical}{Analytical}
	\SetKwInOut{Input}{input}\SetKwInOut{Output}{output}
	\Analytical{$\boldsymbol{\psi}_{initial}, \boldsymbol{\theta}$}
	
	\Input{Current system state, $\boldsymbol{\psi}_{initial}$; schedule, $\boldsymbol{\theta}$; current time, $t_{cur}$}
	\Output{Ratio objective function, $\lambda$; state, $\boldsymbol{\psi}$}
	
	\nl $\zeta \leftarrow 0$, $T_{l} \leftarrow t_{cur}$ \tcp{initialise weighted downtime and start time} \label{line:initialise}
	\nl \While{there are tasks remaining in the schedule, $\boldsymbol{\theta}$}{
		\nl \If{$\boldsymbol{\theta}_{1} = 0$}{ \label{line:sreplenpoint} 
			\nl $T_{a} \leftarrow T_{l} + \frac{s_{ar}}{V_{a}}$ \tcp{calculate arrival time} \label{line:sreplenarrivereplen}
			\nl $T_{f} \leftarrow  T_{a} + D_{sr} + \frac{c_{a} - L_{a}}{R_{r}} + D_{pr}$ \tcp{replenishment finish time} \label{line:sfinishreplenr}
			\nl $L_{a} \leftarrow c_{a}$ \tcp{reset resource level} \label{line:sreplenlevel}
		}
		\nl \Else{
			\nl $i \leftarrow \boldsymbol{\theta}_{1}$ \tcp{index of target user agent}
			\nl $T_{a} \leftarrow T_{l} + \frac{s_{au,i}}{V_{a}}$ \tcp{calculate arrival time} \label{line:sreplenarriveuser}
			\nl $T_{b,i} \leftarrow T_{a} + D_{sa}$ \tcp{time after set-up} \label{line:scalctimeaftersetup}
			\nl $T_{d,i} \leftarrow T_{f,i} + \frac{L_{u,i}}{R_{u,i}}$ \tcp{deadline} \label{line:sdeadline}
			\nl $D_{c,i} \leftarrow T_{b,i}-T_{d,i}$ \tcp{downtime} \label{line:sdowntime0}
			\nl $E(\textrm{downtime}) \leftarrow \int\limits_{0}^{\infty} d \ p(D_{c,i}) \; \textrm{d}d$ \tcp{expected downtime} \label{line:sexpecteddowntime}
			\nl $\zeta \leftarrow \zeta + w_{i}E(\textrm{downtime})$ \tcp{weighted downtime} \label{line:sdowntime1}
			\nl $L_{u,i} \leftarrow \left(L_{u,i} - \left(T_{b,i} - T_{f,i}\right)R_{u,i}\right)^{\#}$ \tcp{level before replenishment} \label{line:suserlevel}
			\nl $Q_{u,i} \leftarrow \left(c_{u,i} - L_{u,i}\right)\frac{R_{a}}{R_{a} - R_{u,i}}$ \label{line:squan} \tcp{replenishment quantity}
			\nl $Q_{u,i}^{*} \leftarrow \left(Q_{u,i}\right)^{*\le L_{a}}$ \label{line:squan2} \tcp{adjusted quantity}
			\nl $D_{r,i} \leftarrow \frac{Q_{u,i}^{*}}{R_{a}}$ \tcp{replenishment duration} \label{line:stimetoreplen}
			\nl $L_{u,i} \leftarrow \left(L_{u,i} - \left(T_{b,i} - T_{f,i}\right)R_{u,i}\right)^{\#}$ \tcp{user agent level} \label{line:snewleveluser}
			\nl $L_{a} \leftarrow  \left(L_{a} - Q_{u,i} \right)^{\#}$ \tcp{replenishment agent level} \label{line:snewlevelreplen}
			\nl $T_{f,i} \leftarrow  T_{b,i} + D_{r,i}$ \tcp{update last replenishment time} \label{line:slastreplen}
			\nl $T_{l} \leftarrow  T_{f,i} + D_{pa}$ \tcp{time after pack-up} \label{line:stimeafterpackup}
		}
		\nl remove the task $\boldsymbol{\theta}_{1}$ from the schedule, $\boldsymbol{\theta}$\; \label{line:removetask}
	}
	\nl \ForAll{user agents, $i$}{ \label{line:final_downtime}
		\nl $T_{d,i} \leftarrow T_{f,i} + \frac{L_{u,i}}{R_{u,i}}$ \tcp{deadline}
		\nl $D_{c,i} \leftarrow T_{l}-T_{d,i}$ \tcp{downtime}
		\nl $E(\textrm{downtime}) \leftarrow \int\limits_{0}^{\infty} d \ p(D_{c,i}) \; \textrm{d}d$ \tcp{expected downtime}
		\nl $\zeta \leftarrow \zeta + w_{i} E(\textrm{downtime})$ \tcp{weighted downtime} \label{line:sdowntime2}
	}
	\nl $\boldsymbol{\psi} \leftarrow \left(L_{a}, L_{u,i} \; \forall i \right)$ \tcp{predicted final state} \label{line:predicted_state}
	\nl $\lambda \leftarrow \frac{\zeta}{n E(T_{l} - t_{cur})}$ \tcp{ratio objective function} \label{line:ratio_objective}
	\caption{Prediction framework} \label{a:single}
\end{algorithm}

On line \ref{line:initialise}, the total weighted downtime is initialised to 0 and the start time of the schedule is initialised to the current time. Note that the start time, $T_{l}$, is initialised as a random variable with no uncertainty. Then, while there are tasks remaining in the schedule, the first task in the schedule is evaluated. If the task is for the replenishment agent to visit the replenishment point (line \ref{line:sreplenpoint}), then the random variable describing the arrival time of the replenishment agent at the replenishment point, $T_{a}$, is calculated as per line \ref{line:sreplenarrivereplen}. The time that the replenishment is completed at, $T_{f}$, is given on line \ref{line:sfinishreplenr}, and the resource level of the replenishment agent is reset to its capacity on line \ref{line:sreplenlevel}. 

If the task is instead to travel to and replenish a user agent, then the arrival time is calculated on line \ref{line:sreplenarriveuser}, and the time after the replenishment agent has set up, $T_{b,i}$, is calculated on line \ref{line:scalctimeaftersetup}. The time at which the user agent would exhaust its supply of the resource if it were not replenished, $T_{d,i}$, is calculated on line \ref{line:sdeadline}, where $T_{f,i}$ is the time that the user agent was last replenished (see line \ref{line:slastreplen}). If the user agent has not been visited by a replenishment agent, then $T_{f,i} = t_{cur}$. The random variable describing the duration of downtime incurred by the user agent between when it was last replenished and the start time of the current replenishment action, $D_{c,i}$, is calculated on line \ref{line:sdowntime0}. 

The random variable $D_{c,i}$ is described by a probability distribution over downtime duration, $d$, which gives the probability of any duration of downtime being incurred. When considering downtime, negative downtime is equivalent to uptime and does not incur a cost in this problem formulation, and, therefore, $D_{c,i}$ is only of interest in the positive domain. The expected downtime of user agent $i$, $E(\textrm{downtime})$, given the probability distribution of downtime, $p(D_{c,i})$, is calculated by the integral on line \ref{line:sexpecteddowntime}. The solution to this integral when using Gaussian distributed random variables is presented in Section \ref{s:expected_value}. The expected downtime is then added to the running total of weighed downtime on line \ref{line:sdowntime1}. 

The resource level of the user agent before the replenishment begins, $L_{u,i}$, is calculated on line \ref{line:suserlevel}, where the \# operator denotes that the distribution has been adjusted to account for hard constraints on the state of the system. Note that this adjustment is only required when using probability distributions that do not already fit within the hard constraints (i.e. probability distributions that have infinite domain). A method for adjusting Gaussian distributions against hard constraints is discussed in Section \ref{s:adjusting_hard}. The quantity of the resource required to fully replenish the user agent, $Q_{u,i}$, is given on lines \ref{line:squan} and \ref{line:squan2}. Here, $*\le L_{a}$ means that $Q_{u,i}$ is adjusted so that it does not exceed the resource level of the replenishment agent, $L_{a}$, as the replenishment agent may have insufficient supply of the resource to fully replenish the user agent. A method for performing this soft adjustment for Gaussian distributions is outlined in Section \ref{s:soft}. 

The time taken to replenish the user agent is then given on line \ref{line:stimetoreplen}. The new levels of the user agent and replenishment agent after the replenishment are calculated on lines \ref{line:snewleveluser} and \ref{line:snewlevelreplen} respectively. Note that the unadjusted $Q_{u,i}$ is used on line \ref{line:snewlevelreplen} as using $Q_{u,i}^{*}$ can underestimate the amount of the resource transferred by the replenishment agent. The $\#$ adjustment ensures that $L_{a}$ remains non-negative. The time that the user agent was last replenished, $T_{f,i}$, is updated on line \ref{line:slastreplen}, and the time that the replenishment agent finishes packing up, $T_{l}$, is given on line \ref{line:stimeafterpackup}. The task is then removed from the schedule on line \ref{line:removetask}. 

The final block of the algorithm (lines \ref{line:final_downtime}-\ref{line:sdowntime2}) calculates whether any of the user agents incur additional downtime between when they are last replenished and the completion time of the schedule. The algorithm concludes by returning the predicted resource levels of the agents (line \ref{line:predicted_state}) and the ratio objective function (line \ref{line:ratio_objective}). The duration of the schedule, given by $T_{l} - t_{cur}$, is a random variable, and the expected value used on line (\ref{line:ratio_objective}) is simply the mean value of the probability distribution describing $T_{l} - t_{cur}$. 

\section{Using Gaussian distributed random variables} \label{s:gaussian}

Analytical methods exist for calculating the sum of Gaussian distributed random variables, but do not exist for multiplication, division, and other operations used in the prediction framework presented in the previous section. To enable the use of Gaussian distributed random variables, this section introduces approximations to the inverse of a Gaussian distributed random variable, ratio of Gaussian distributed random variables, and product of Gaussian distributed random variables, as well as methods for evaluating the expected downtime, and adjusting Gaussian distributed random variables against hard and soft constraints. 

\subsection{Inverse Gaussian distributed random variable}\label{s:inverse}

A Gaussian approximation to the inverse of a Gaussian distributed random variable was presented in \cite{Palmer2013}. A new Gaussian approximation is introduced here which will be shown in Section \ref{s:compstud} to outperform the one presented in \cite{Palmer2013}. Consider an inverse Gaussian distributed variable, $I$, that is formed by:

\begin{equation}
I = \frac{c}{G}
\end{equation}
where $G$ is a Gaussian distributed variable with mean $u_{G}$ and standard deviation $\sigma_{G}$, and $c$ is a constant. A Gaussian approximation of $I$ can be attained by assuming that the points at $\mu_{G} + \sigma_{G}$ and $\mu_{G} - \sigma_{G}$ give the equivalent points at $\mu_{I} - \sigma_{I}$ and $\mu_{I} + \sigma_{I}$ respectively, where $\mu_{I}$ and $\sigma_{I}$ are the mean and standard deviation of the Gaussian approximation of $I$. These give the following values for $\mu_{I}$ and $\sigma_{I}$:

\begin{align}
\mu_{I} &= \frac{1}{2}\left(\frac{c}{\mu_{G} - \sigma_{G}} + \frac{c}{\mu_{G} + \sigma_{G}}\right)\notag \\
& = \frac{c\mu_{G}}{\mu_{G}^2 - \sigma_{G}^{2}} \label{eq:inv_mean}
\end{align}

\begin{align}
\sigma_{I} &= \frac{1}{2}\left(\frac{c}{\mu_{G} - \sigma_{G}} - \frac{c}{\mu_{G} + \sigma_{G}}\right)\notag \\
& = \frac{c\sigma_{G}}{\mu_{G}^2 - \sigma_{G}^{2}}\label{eq:inv_std_dev}
\end{align}

\begin{figure}
	\centering
	\includegraphics[width=0.6\textwidth]{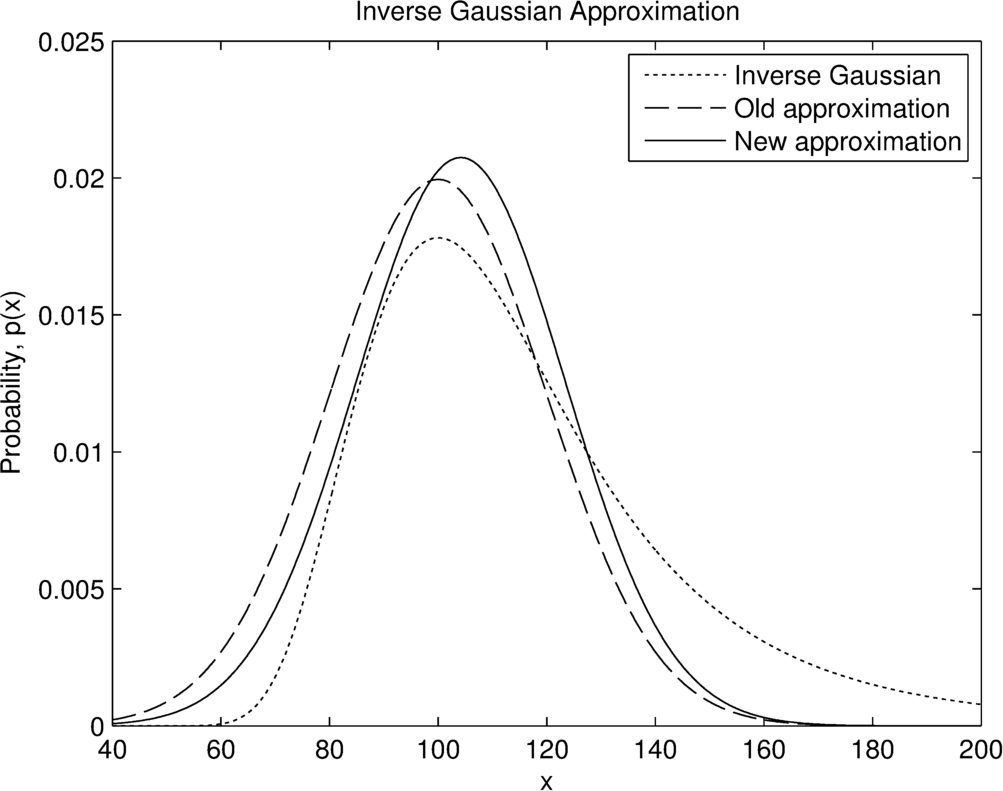}
	\caption{A comparison of the approximations of the inverse Gaussian distribution for $\mu_{G}/\sigma_{G} = 5$. }
	\label{f:inverse}
\end{figure}

When $\sigma_{G}$ is small in comparison to $\mu_{G}$, the inverse Gaussian distributed variable is highly Gaussian in shape. As $\sigma_{G}$ is increased, the resultant inverse Gaussian distributed variable is skewed further to the right. The advantage of the approximation developed here over the one previously presented in \cite{Palmer2013} is demonstrated in Figure \ref{f:inverse} for a case with high uncertainty. As can be seen, the new approximation better approximates the inverse Gaussian distribution, particularly in the left tail where the old approximation significantly overestimates the probability. 

\subsection{Ratio of Gaussian distributed variables}

A method for approximating the ratio of two Gaussian distributed variables is given in \cite{Marsaglia2006}. A ratio, $R$:

\begin{equation}
R = \frac{E}{F}
\end{equation}
with $E \sim \mathcal{N}\left(\mu_{E}, \sigma_{E}\right)$ and $F \sim \mathcal{N}\left(\mu_{F}, \sigma_{F}\right)$, and correlation between $E$ and $F$ of $\rho = 0$, can be approximated with a Gaussian distributed random variable where:

\begin{equation}
r = \frac{\sigma_{F}}{\sigma_{E}}, a =  \frac{\mu_{E}}{\sigma_{E}} \textrm{ and } b =  \frac{\mu_{F}}{\sigma_{F}}
\end{equation}
\begin{equation} \label{eq:ratiomu}
\mu_{R} = \frac{a}{r(1.01b - 0.2713)}
\end{equation}
\begin{equation} \label{eq:ratiosigma}
\sigma_{R} = \frac{1}{r}\sqrt{\frac{a^{2}+1}{b^{2} + 0.108b - 3.795} - r^{2}\mu_{R}^{2}}
\end{equation}

The authors specified that the approximation is only valid for $a < 2.5$, $b > 4$ \cite{Marsaglia2006}. As $a \rightarrow \infty$, the ratio of Gaussian distributed random variables is similar to an inverse Gaussian distributed random variable and can be approximated using the inverse Gaussian approximation presented above. For situations where $a \ge 2.5$, the inverse Gaussian approximation method with $E$ treated as a scalar, $e = \mu_{E}$, has been used. 

\subsection{Product of Gaussian distributed variables}

An approximation to the product of two Gaussian distributed variables is presented in \cite{Ware2003}. For a product, $M$:

\begin{equation}
M = EF
\end{equation}

Then:

\begin{equation}
\mu_{M} = \mu_{E}\mu_{F}
\end{equation}

\begin{equation}
\sigma_{M}^{2} = \sigma_{E}^{2}\sigma_{F}^{2}(1 + \delta_{E}^{2} + \delta_{F}^{2})
\end{equation}
where

\begin{equation}
\delta_{x} = \frac{\mu_{x}}{\sigma_{x}}
\end{equation}

The authors noted that the approximation improves as $\delta_{E}$ and $\delta_{F}$ become large \cite{Ware2003}. 

\subsection{Expected value} \label{s:expected_value}

The integral required for calculating the risk-weighted downtime on line \ref{line:sdowntime0} of Algorithm \ref{a:single} when using Gaussian distributions equates to:

\begin{align}
E(D_{c,i}) & = \int\limits_{0}^{\infty } \! d \ p(D_{c,i}) \ \textrm{d}d \notag \\
&= \int\limits_{0}^{\infty } \! \frac{d}{\sigma\sqrt{2\pi}} \exp \left(\frac{-(d-\mu)^{2}}{2\sigma^2}\right) \ \textrm{d}d \notag\\
&= \frac{\mu}{2}\left(1+\textrm{erf}\left(\frac{\mu}{\sigma\sqrt{2}}\right)\right) + \frac{\sigma}{\sqrt{2\pi}}\exp \left(-\frac{\mu^{2}}{2\sigma^{2}}\right)
\end{align}
where $\mu$ and $\sigma$ are the mean and standard deviation of the probability distribution describing the random variable $D_{c,i}$. 

\subsection{Adjusting against hard constraints} \label{s:adjusting_hard}

The probability distributions describing several random variables were required to be adjusted to take into consideration hard limitations on the system state; for example, the resource level of a user agent is bounded by 0 and $c_{u,i}$. A novel Gaussian approximation to the generalised rectified Gaussian distribution is introduced here. The rectified Gaussian distribution, used by \cite{Meng2011}, groups the probability in the negative domain at 0. The generalised rectified Gaussian distribution is proposed as an extension to this, where the distribution is rectified between two arbitrary values, $a$ and $b$. If the original CDF of the Gaussian distribution is $F(x)$, the problem is to calculate a new Gaussian PDF, $\mathcal{N}(\mu_{R},\sigma_{R}^{2})$, that approximates the PDF of the generalised rectified Gaussian distribution that satisfies the following CDF, $F_{R}(x)$:
\begin{equation}
F_{R}(x) = \begin{cases}
0 & \textrm{if } x < a\\
F(x)  & \textrm{if } a \le x < b \\
1 & \textrm{if } x \ge b\\
\end{cases}
\end{equation}
where $a$ and $b$ are the limits on the state. The following process is similar to the truncation approach for constrained Kalman filtering \cite{Palmer2016}. The distribution being adjusted is first transformed to a standard normal distribution, yielding transformed constraints of $c$ and $d$ respectively:
\begin{equation}
c = \frac{a - \mu_{A}}{\sigma_{A}} \qquad d = \frac{b - \mu_{A}}{\sigma_{A}}
\end{equation}
where $\mu_{A}$ and $\sigma_{A}$ are the mean and standard deviation of the distribution being adjusted. The mean and variance of the Gaussian approximation of the rectified Gaussian distribution are then given by:
\begin{equation}\label{eq:adjust_truncated_mu}
\begin{aligned}
\mu_{z} &= \int\limits_{c}^{d} \frac{\zeta}{\sqrt{2\pi}} \exp \left(-\frac{\zeta^{2}}{2} \right) \textrm{d}\zeta  + \frac{c}{2}\left(1 + \textrm{erf}\left( \frac{c}{\sqrt{2}}\right) \right) \\
& \qquad + \frac{d}{2}\left(1 - \textrm{erf}\left( \frac{d}{\sqrt{2}}\right) \right)\\
& = \frac{1}{\sqrt{2\pi}} \left(\exp \left( \frac{-c^{2}}{2}\right) - \exp \left( \frac{-d^{2}}{2}\right)\right) \\
& \qquad + \frac{c}{2}\left(1 + \textrm{erf}\left( \frac{c}{\sqrt{2}}\right) \right) + \frac{d}{2}\left(1 - \textrm{erf}\left( \frac{d}{\sqrt{2}}\right) \right)
\end{aligned}
\end{equation}
\begin{equation}
\begin{aligned}
&\sigma_{z}^{2} = \int\limits_{c}^{d} \left(\zeta - \mu \right)^{2} \exp \left(-\frac{\zeta^{2}}{2} \right) \textrm{d}\zeta \\
& \qquad + \frac{\left(c-\mu\right)^{2}}{2}\left(1 + \textrm{erf}\left( \frac{c}{\sqrt{2}}\right) \right) \\
& \qquad + \frac{\left(d-\mu\right)^{2}}{2}\left(1 - \textrm{erf}\left( \frac{d}{\sqrt{2}}\right) \right)\\
& = \frac{\mu^{2} + 1}{2}\left(\textrm{erf}\left(\frac{d}{\sqrt{2}}\right) - \textrm{erf}\left(\frac{c}{\sqrt{2}}\right) \right) \\
&  - \frac{1}{\sqrt{2\pi}}\left(\exp\left(-\frac{d^{2}}{2}\right)\left(d-2\mu\right) - \exp\left(-\frac{c^{2}}{2}\right)\left(c-2\mu\right)\right)  \\
&  + \frac{\left(c - \mu\right)^{2}}{2}\left(1 + \textrm{erf}\left(\frac{c}{\sqrt{2}}\right)\right) \\
& + \frac{\left(d - \mu\right)^{2}}{2}\left(1 - \textrm{erf}\left(\frac{d}{\sqrt{2}}\right)\right)
\end{aligned}
\end{equation}

Taking the inverse of the transformation gives:

\begin{equation}
\mu_{R} = \mu_{z}\sigma_{A} + \mu_{A} \qquad \sigma^{2}_{R} = \sigma^{2}_{z}\sigma^{2}_{A}
\end{equation}

\subsection{Adjusting against soft constraints}\label{s:soft}

The other type of adjustment used adjusts one random variable so that it does not exceed another random variable. This soft adjustment is denoted by a * followed by the variable that it is adjusted against. This is used on line \ref{line:squan2} of Algorithm \ref{a:single} to ensure that the quantity of the resource used to replenish a user agent does not exceed the current capacity of the replenishment agent. Consider a random variable, $A$, that is adjusted so that it does not exceed the random variable $B$. The proposed method is as follows:

\begin{equation}
\mu_{A}^{*\le B} = \begin{cases}
\mu_{A} & \textrm{if } \mu_{A} - 3\sigma_{A} < \mu_{B} - 3\sigma_{B} \\
& \textrm{and } \mu_{A} + 3\sigma_{A} < \mu_{B} + 3\sigma_{B} \\ 
\mu_{B} & \textrm{if } \mu_{A} - 3\sigma_{A} > \mu_{B} - 3\sigma_{B} \\
& \textrm{and } \mu_{A} + 3\sigma_{A} > \mu_{B} + 3\sigma_{B} \\ 
\frac{\mu_{B} - 3\sigma_{B} + \mu_{A} + 3\sigma_{A}}{2} & \textrm{if } \mu_{A} - 3\sigma_{A} > \mu_{B} - 3\sigma_{B}  \\
& \textrm{and } \mu_{A} + 3\sigma_{A} < \mu_{B} + 3\sigma_{B} \\ 
\frac{\mu_{A} - 3\sigma_{A} + \mu_{B} + 3\sigma_{B}}{2} & \textrm{if } \mu_{A} - 3\sigma_{A} < \mu_{B} - 3\sigma_{B} \\
& \textrm{and } \mu_{A} + 3\sigma_{A} > \mu_{B} + 3\sigma_{B}
\end{cases}
\end{equation}

\begin{equation}
\sigma_{A}^{*\le B} = \begin{cases}
\sigma_{A} & \textrm{if } \mu_{A} - 3\sigma_{A} < \mu_{B} - 3\sigma_{B}  \\
& \textrm{and } \mu_{A} + 3\sigma_{A} < \mu_{B} + 3\sigma_{B} \\ 
\sigma_{B} & \textrm{if } \mu_{A} - 3\sigma_{A} > \mu_{B} - 3\sigma_{B}\\
& \textrm{and } \mu_{A} + 3\sigma_{A} > \mu_{B} + 3\sigma_{B} \\ 
\frac{\mu_{A} + 3\sigma_{A} - \left(\mu_{B} - 3\sigma_{B}\right)}{6} & \textrm{if } \mu_{A} - 3\sigma_{A} > \mu_{B} - 3\sigma_{B} \\
& \textrm{and } \mu_{A} + 3\sigma_{A} < \mu_{B} + 3\sigma_{B} \\ 
\frac{\mu_{B} + 3\sigma_{B} - \left(\mu_{A} - 3\sigma_{A}\right)}{6} & \textrm{if } \mu_{A} - 3\sigma_{A} < \mu_{B} - 3\sigma_{B} \\
& \textrm{and }\mu_{A} + 3\sigma_{A} > \mu_{B} + 3\sigma_{B}
\end{cases}
\end{equation}

This method ensures that $P(A \le x) \le P(B \le x)$ for $x$ within 3 standard deviations of the mean of both $A$ and $B$.

\section{Optimisation methods} \label{s:optmeth}

This section presents the Apparent Tardiness Cost (ATC) heuristic in Section \ref{s:heur}, a simulated annealing meta-heuristic in Section \ref{s:meta}, and a branch and bound method in Section \ref{s:anytime}. All of these methods are used within an MPC-like framework---after each task is performed, the optimisation method is rerun to calculate a new task to be performed. This replanning enables unexpected changes to the state of the system to be considered by the optimisation. Before the optimisation method is run, the resource level of the replenishment agent is first checked to see whether it is above a threshold, $l_{a,thresh}$. If the resource level is below the threshold, then the replenishment agent is immediately sent back to the replenishment point to be replenished. A threshold of 5\% of maximum capacity was found to give good results in the scenarios considered in this paper. 

The ATC heuristic calculates what the next task of the replenishment agent should be, while the simulated annealing and branch and bound approaches both consider a schedule of tasks. The finite horizon used in this paper is the number of tasks in the schedule. This is to enable fair comparison between the simulated annealing and branch and bound approaches. 

\subsection{ATC heuristic} \label{s:heur}

The ATC heuristic used by Kaplan and Rabadi in \cite{Kaplan2012} for the aerial refuelling problem is a combination of the Weighted Shortest Processing Time first (WSPT) and Minimum Slack first (MS) rules. It calculates priorities for each task based on the following formula:

\begin{equation}\label{eq:atc}
\pi_{i} = \frac{w_{i}}{d_{l,i}}\phi_{i}
\end{equation}
where $\pi_{i}$ is the priority of task $i$ determined by the heuristic, $d_{l,i}$ is the total duration of the task, and $\phi_{i}$ is the marginal cost of delay. The task with the highest priority is selected as the next task to be performed. The marginal cost of delay used in \cite{Kaplan2012} combined soft and hard deadlines with a ready time for each task. The SCAR scenarios under consideration only have a soft deadline, yielding a marginal cost of delay of:

\begin{equation}
\phi_{i} = \exp\left(-\frac{\max(0,t_{d,i} - t_{b,i})}{k\overline{t_{b}}}\right)
\end{equation}
where $t_{b,i}$ is the time at which the replenishment agent begins replenishing the user agent, $\overline{t_{b}}$ is the average start time for all possible replenishment tasks for that replenishment agent, $t_{d,i}$ is the deadline for the user agent, and $k$ is a scaling factor. The scaling factor biases the behaviour of the ATC heuristic towards the WSPT rule if $k$ is very large, and towards the MS rule if $k$ is very small. Typical values of $k$ used range between 1 and 7. It should be noted that the deadline used by \cite{Kaplan2012} is the time by which the task must be completed, whereas the deadline in a SCAR scenario is the time before which the replenishment agent must begin replenishing the user agent. The deadline is calculated for each user agent $i$ as:

\begin{equation} \label{eq:timeempty}
t_{d,i} = \frac{l_{u,i}}{r_{u,i}}
\end{equation}
where $r_{u,i}$ is the mean value of $R_{u,i}$. The start time for replenishing each user agent $i$ is calculated as:

\begin{equation}
t_{b,i} = \frac{s_{au,i}}{v_{a}} + d_{sa}
\end{equation}
where $v_{a}$ is the mean value of $V_{a}$, and $d_{sa}$ is the mean value of $D_{sa}$. The total duration for each user agent, $d_{l,i}$, is given by:

\begin{equation}
d_{l,i} = t_{b,i} + \frac{c_{u,i} - \max\left(0, \; l_{u,i} - r_{u,i}t_{b,i}\right)}{r_{a}- r_{u,i}} + d_{pa}
\end{equation}
where $r_{u,i}$ is the mean value of $R_{u,i}$, $r_{a}$ is the mean value of $R_{a}$, and $d_{pa}$ is the mean value of $D_{pa}$. 

\subsection{Simulated annealing} \label{s:meta}

The simulated annealing method used by Kaplan and Rabadi \cite{Kaplan2012, Kaplan2013} was implemented, using the ATC heuristic to generate an initial schedule. The algorithm moves to neighbour solutions by randomly replacing a task in the schedule with one of the other possible tasks. The following limitation was placed on the generated schedule, $\boldsymbol{\theta}$:

\begin{equation}
\boldsymbol{\theta}_{i-1} \ne \boldsymbol{\theta}_{i} \ne \boldsymbol{\theta}_{i+1} \quad \forall i \in \{2,3 ... a-1\}
\end{equation}
where $a$ is the number of tasks in the schedule. This ensures that successive tasks are different. The inputs to the simulated annealing algorithm are an initial temperature coefficient, a temperature cooling coefficient, a maximum number of inner loop iterations, and a maximum number of iterations. Values for these parameters are suggested in \cite{Kaplan2012}. For the two scenarios examined in Section \ref{s:compstud}, a maximum number of inner loop iterations of 25, and a maximum number of iterations of 2000 worked well. For Scenario 1, an initial temperature coefficient of 0.2, and a temperature cooling coefficient of 0.95 gave good results, while in Scenario 2 an initial temperature coefficient of 0.1, and a temperature cooling coefficient of 0.7 performed well. The cost of each schedule is evaluated using the prediction framework from Section \ref{s:pred}. 

\subsection{Branch and bound} \label{s:anytime}

The set of all possible schedules for a given finite horizon forms a tree where each branch represents the possible choices for the next task. Branch and bound minimises the size of the state space that is explored by culling branches of the tree where the minimum possible cost is higher than the cost of the best solution found so far \cite{Land1960}. Similar to the simulated annealing implementation, the branch and bound implementation restricts consecutive tasks to be different. In addition, if the resource level of the replenishment agent is below the threshold $l_{a,thresh}$ at any point in the tree, the only valid task to be performed next is for the replenishment agent to be replenished by the replenishment point. 

An important aspect of branch and bound is estimating the lower bound on the cost for each node. This lower bound represents the lowest possible cost of a complete schedule starting with the sequence of tasks described by that node. The more accurate the estimate of the lower bound, the more branches that can be pruned. A heuristic for estimating the minimum possible cost from any node in the tree for a SCAR scenario was developed in the authors' previous work \cite{Palmer2014a} and has been used in this paper. Essentially, it assumes that the user agents do not exhaust their supply of the resource, while also providing a conservative estimate of the total time of the schedule. This guarantees that the estimated minimum cost of a schedule is below the actual cost. 

\begin{figure}
	\centering
	\includegraphics[width=0.8\textwidth]{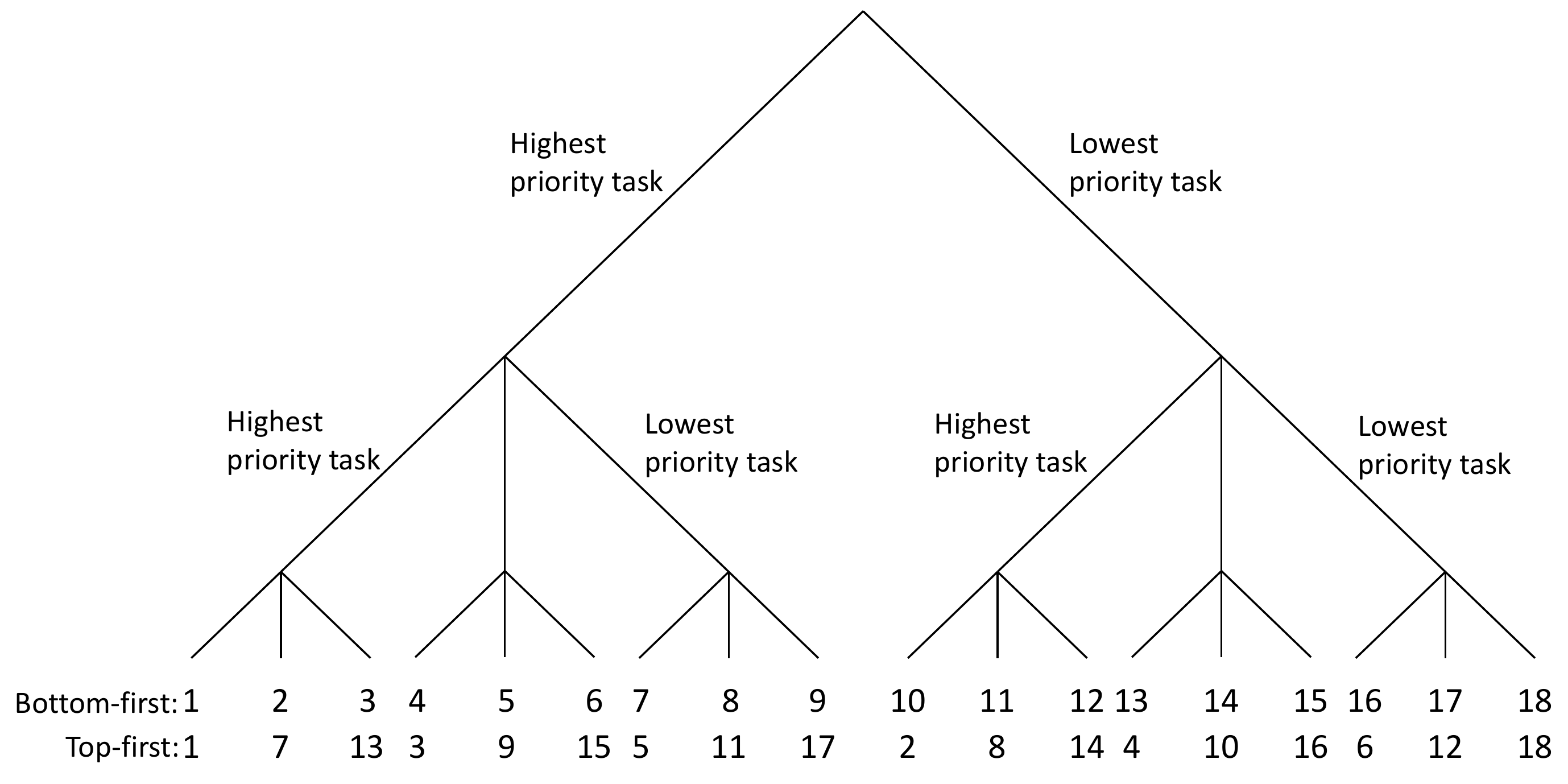}
	\caption{Two different methods for searching through a tree. The top line shows the exploration order using the bottom-first method, while the bottom line shows the exploration order using the top-first method.}
	\label{f:methods}
\end{figure}

To improve the search speed of the algorithm, the ATC heuristic was used to generate priorities for the tasks branching from each node. Two different methods of exploring the tree, shown in Figure \ref{f:methods}, were considered. The first method, bottom-first, involves searching through the leaves first, and then gradually searching higher in the tree. This method has the advantage of not requiring any data to be stored in a tree structure, but has the disadvantage of focussing on one branch initially. The other method, top-first, explores the tree using a top down approach---the nodes are explored in priority order with changes initially occurring at the top level. As can be seen, each successive solution examined is in the opposite high level branch to the previous solution, ensuring that the breadth of the tree is explored rapidly. However, the top-first method requires the calculated costs and lower bounds of every node visited to be stored in a tree, which results in a memory complexity of O($n^{a}$), where n is the number of user agents and a is the number of tasks in the schedule. 

\begin{figure}
	\centering
	\includegraphics[width=0.6\textwidth]{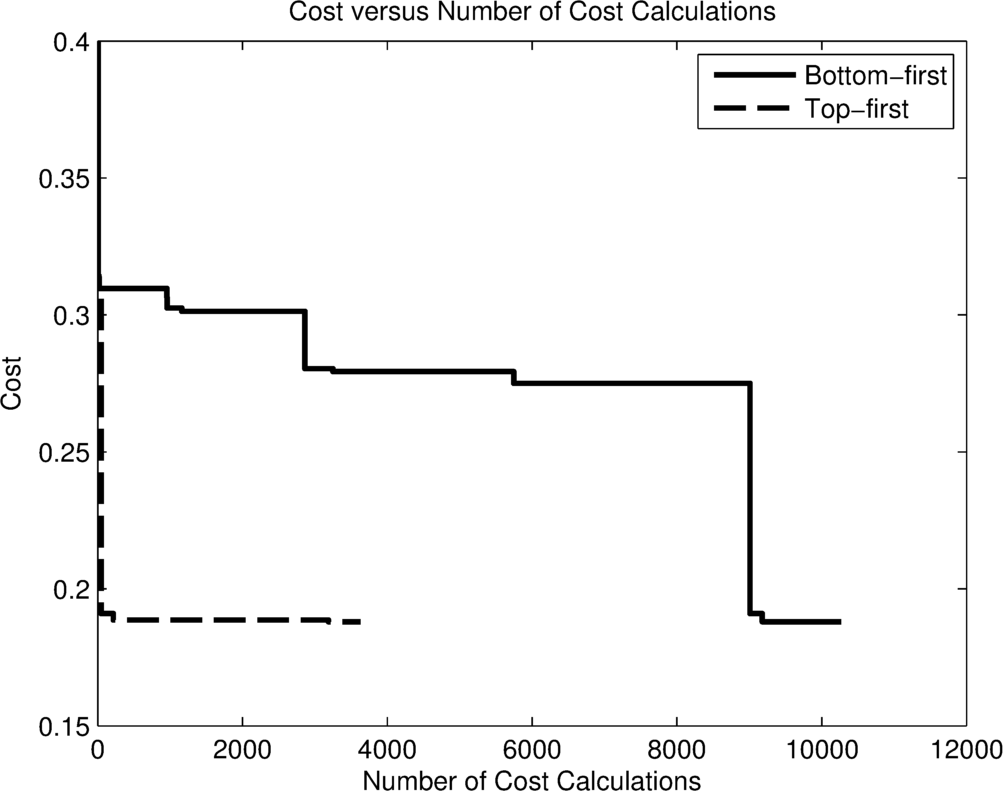}
	\caption{Cost versus number of calculations for the two branch and bound exploration methods. Lower costs are better. The bottom-first method incrementally improves on the solution, while the top-first method very quickly finds better solutions in other branches of the tree. In this example, the bottom-first method found the optimal schedule in 9178 calculations and required a total of 10269 calculations to fully explore the tree. The top-first method found the optimal schedule in 3186 calculations and required only 3629 calculations in total. }
	\label{f:costcalcs}
\end{figure}

The cost of the current best schedule versus the number of nodes explored for the two methods is compared in Figure \ref{f:costcalcs} for a sample scenario. Both methods initially examine the same schedule generated by the ATC heuristic before searching other areas of the tree. Where the bottom-first approach finds neighbour schedules which make minor incremental improvements to the cost, the top-first approach quickly finds substantially better schedules in other branches of the tree. The bottom-first approach has many desirable characteristics for small optimisation problems---low memory usage and minimal computational overhead associated with having to search through the tree. In larger problems, however, it may be computationally intractable to search through the entire tree and the anytime characteristic of branch and bound must be exploited. In these cases, the top-first approach is more desirable as it generally finds lower cost schedules for the same number of nodes explored as the bottom-first approach. In addition, since it focusses on earlier tasks, it fits quite well into the MPC-like framework used in this paper---optimisation efforts are focussed on the next tasks to be performed rather than the tasks at the end of the schedule. The top-first approach is used for the remainder of this paper. 

\begin{figure}
	\centering
	\includegraphics[width=0.8\textwidth]{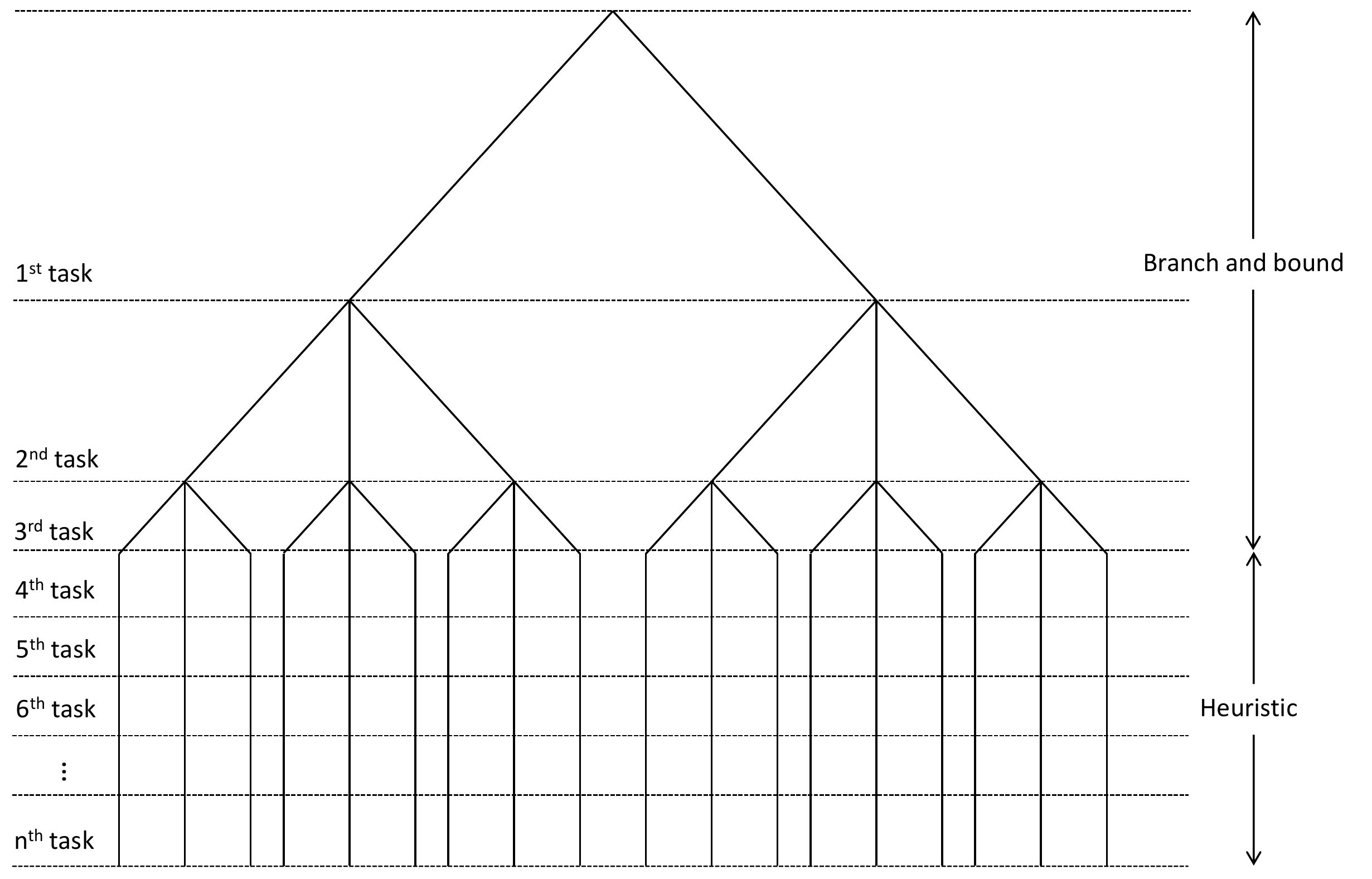}
	\caption{Branch and bound when the optimisation depth is smaller than the schedule length---the tasks within the optimisation depth (3 tasks in this example) are optimised using branch and bound, while the remaining tasks are selected using the ATC heuristic. }
	\label{f:optdepth}
\end{figure}

To take advantage of the anytime characteristic of branch and bound, an optimisation depth was specified. As shown in Figure \ref{f:optdepth}, the optimisation depth determines how far through the tree the branch and bound searches before selecting the remaining tasks using the ATC heuristic. This enables the computation time of the algorithm to be restricted while still using a long finite horizon. If the optimisation depth is equal to the schedule length, branch and bound will return the optimal schedule for that schedule length. 

\section{Computational study} \label{s:compstud}

This section evaluates the prediction and optimisation methods developed in this paper. All methods were implemented by the authors in Python on a 2.8GHz Intel i7-640M. Section \ref{s:scenarios} first introduces the two scenarios used to evaluate the methods. Section \ref{s:comp_pred} then evaluates the prediction method developed in Section \ref{s:pred}. Finally, Section \ref{s:opt_pred} compares the developed branch and bound approach with the existing ATC heuristic and simulated annealing approaches.

\subsection{Scenarios}\label{s:scenarios}

\subsubsection{Scenario 1} \label{s:scenario1}

\begin{figure}
	\centering
	\includegraphics[width=0.6\textwidth]{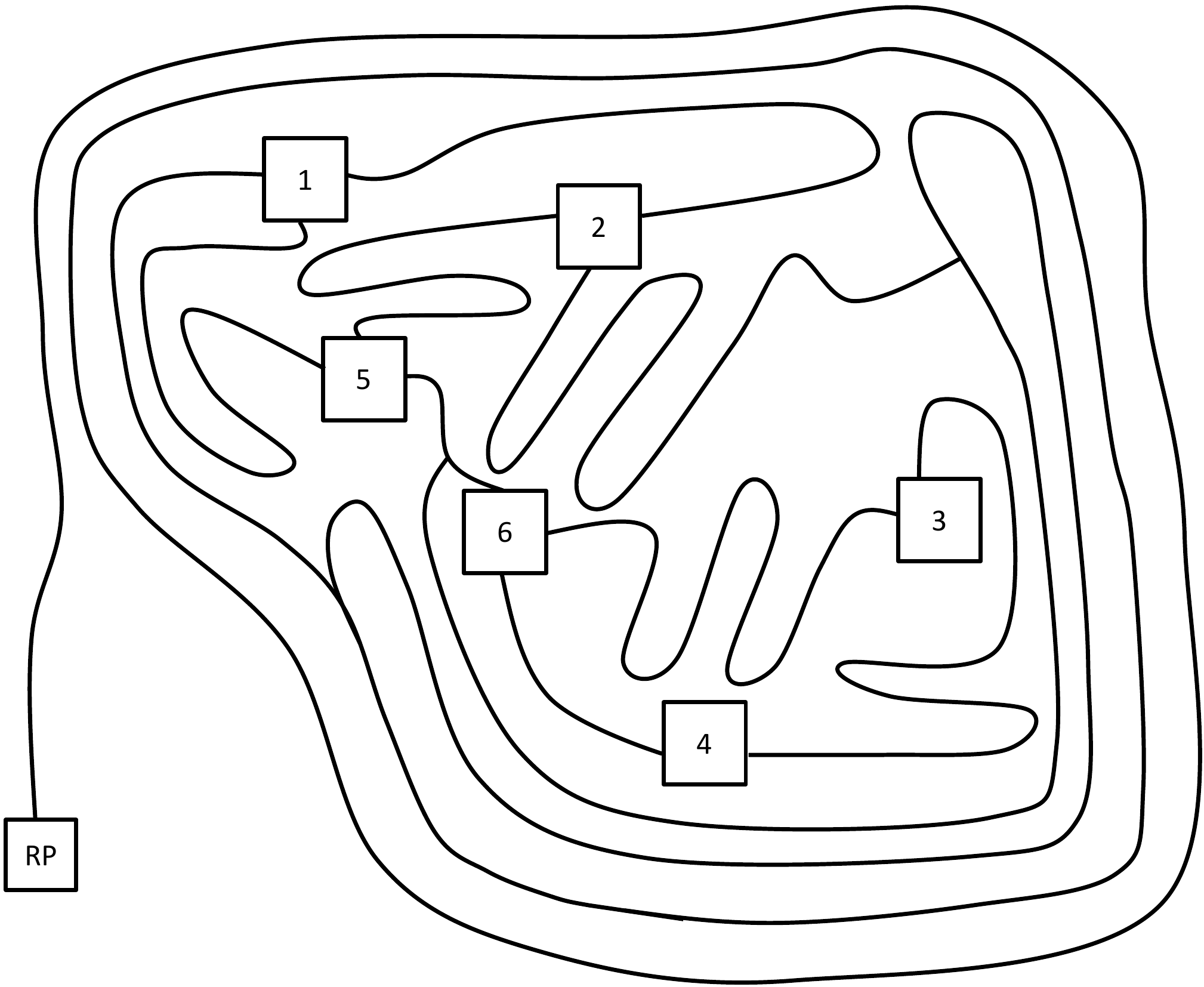}
	\caption{Example mine layout and road network used for Scenario 1. The locations of the Replenishment Point (RP) and the operational areas of the user agents (numbers correspond to the index of the user agent) are indicated by the squares.}
	\label{f:layout_complex}
\end{figure}

\begin{figure}
	\centering
	\includegraphics[width=\textwidth]{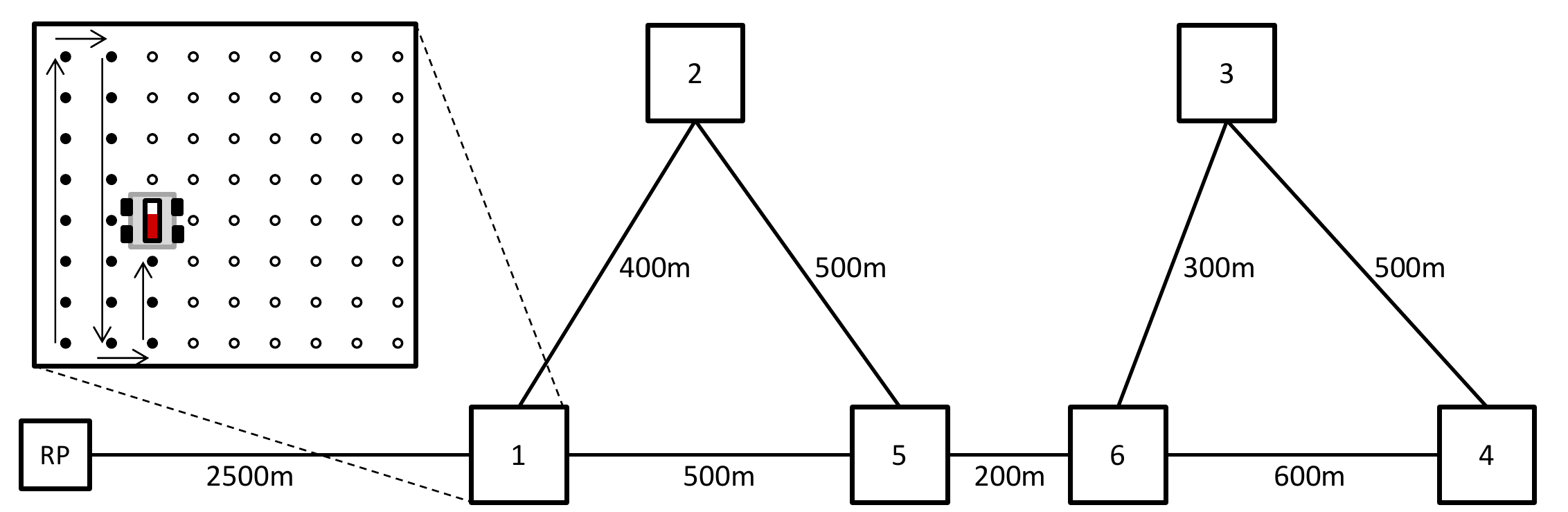}
	\caption{Locations of the Replenishment Point (RP), the operational areas of the user agents (numbers correspond to the index of the user agent), and the simplified graph for Scenario 1. Inset: Drill hole pattern on a bench. Solid circles represent holes that have been drilled, and empty circles are yet to be drilled. While drilling, the drill is stationary at the hole location. }
	\label{f:layout}
\end{figure}

This scenario consists of several drills and excavators operating on specific benches within a mine site. The network of roads connecting the benches and replenishment point is shown in Figure \ref{f:layout_complex}. This road network is first reduced to a simpler graph without affecting the transit times by removing edges that do not form part of the shortest path between any two operational areas or the replenishment point. This simplified graph is shown in Figure \ref{f:layout}, with edge costs representing the distance between nodes. Note that the replenishment agent can pass through the operational area of a user agent without replenishing the user agent. The inset in Figure \ref{f:layout} shows the expected motion of a user agent. In this case, it is a drill that is drilling a specified hole pattern. While it is drilling a hole, it is stationary. Moving between holes is a very small proportion of the operating time of the drill. 

The parameters of the user agents are shown in Table \ref{t:param_u}. The replenishment agent parameters are shown in Table \ref{t:param_r}, and the parameters of the replenishment point are shown in Table \ref{t:param_p}. The scenario was tested using the first 4, 5, and 6 user agents representing an under-utilised, fully-utilised, and over-utilised scenario respectively. In the under-utilised scenario the replenishment agent is operating below its capacity and should be able to prevent all of the user agents from exhausting their supply of the resource, while in the over-utilised scenario the replenishment agent is operating above its capacity. The fully-utilised case sits between the other two. 

\begin{table}
	\caption{User Agent Parameters for Scenario 1}
	\label{t:param_u}
	\centering
	\begin{tabular}{ c  c  c  c  c  c  c}
		\toprule
		Agent & 1 & 2 & 3 & 4 & 5 & 6 \\
		\midrule
		$c_{u}$ (L) & 1000 & 1200 & 700 & 1200 & 1000 & 800 \\
		$R_{u}$ mean (L/s) & 0.5 & 0.5 & 0.3 & 0.5 & 0.4 & 0.4 \\
		$R_{u}$ standard & 0.05 & 0.05 & 0.05 & 0.02 & 0.08 & 0.04 \\
		deviation (L/s) & & & & & & \\
		\bottomrule
	\end{tabular}
\end{table}

\begin{table}
	\caption{Replenishment Agent Parameters for Scenario 1}
	\label{t:param_r}
	\centering
	\begin{tabular}{ c  c  c }
		\toprule
		Parameter & Mean & Standard Deviation \\
		\midrule
		$c_{a}$ (L) & 5000 & -- \\
		$R_{a}$ (L/s) & 10 & 0.5 \\
		$D_{sa}$ (s) & 60 & 20 \\
		$D_{pa}$ (s) & 20 & 5 \\
		$V_{a}$ (m/s) & 15 & 0.5 \\
		\bottomrule
	\end{tabular}
\end{table}

\begin{table}
	\caption{Replenishment Point Parameters for Scenario 1}
	\label{t:param_p}
	\centering
	\begin{tabular}{ c  c  c }
		\toprule
		Parameter & Mean & Standard Deviation \\
		\midrule
		$D_{sr}$ (s) & 30 & 10 \\ 
		$D_{pr}$ (s) & 10 & 1 \\ 
		$R_{r}$ (L/s) & 20 & 1 \\ 
		\bottomrule
	\end{tabular}
\end{table}

\subsubsection{Scenario 2}

\begin{figure}
	\centering
	\includegraphics[width=0.6\textwidth]{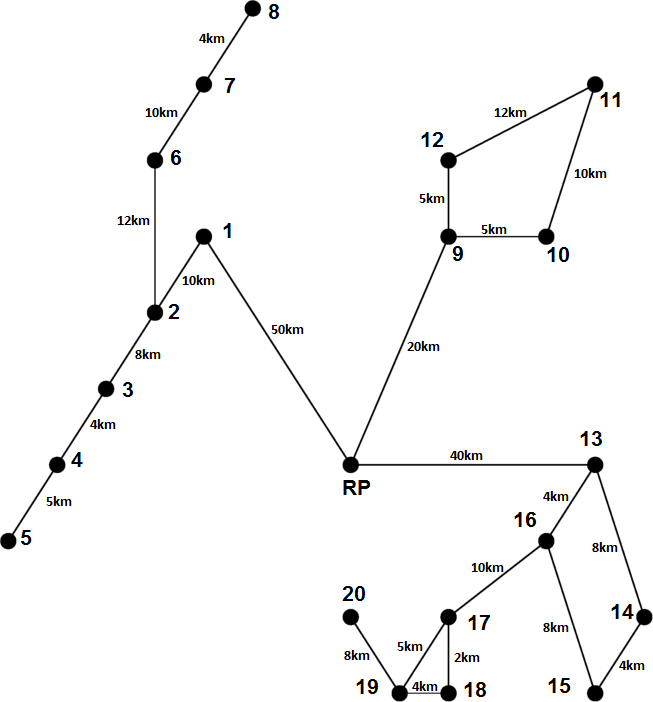}
	\caption{Layout of the 20 user agents in Scenario 2. Road distances are shown in km.}
	\label{f:scen2}
\end{figure}

The second scenario involves the delivery of fuel to 20 agents by truck. This number of agents is representative of large scenarios in the mining and agricultural domains. The quality of roads between operational areas is highly variable which results in travel speeds that are very uncertain. The agents use the fuel at a relatively predictable rate in comparison to the uncertainty of the speed of the truck. Figure \ref{f:scen2} shows the operational areas of the user agents and the simplified graph of the roads connecting them. The full road network has been omitted in the interest of space. 

Three different size replenishment agents were tested, and their parameters are shown in Table \ref{t:param_r2}. The various sizes of the replenishment agent only vary in their capacity---the resource usage rates, set-up and pack-up times, and velocity, are the same for all sizes. The large, medium, and small replenishment agent sizes roughly correspond to under-, fully\nobreakdash-, and over-utilised scenarios respectively. The parameters of the replenishment point are shown in Table \ref{t:param_p2}. There are four different sizes of user agents, each with 2 days supply of fuel. The parameters of each size of user agent are shown in Table \ref{t:param_u2}, and the size of each user agent is outlined in Table \ref{t:param_u3}. 

\begin{table}
	\caption{Replenishment Agent Parameters for Scenario 2}
	\label{t:param_r2}
	\centering
	\begin{tabular}{ c  c  c }
		\toprule
		Parameter & Mean & Standard Deviation \\
		\midrule
		$c_{a}$ Large (L) & 2760 & -- \\
		$c_{a}$ Medium (L) & 1800 & -- \\
		$c_{a}$ Small (L) & 1200 & -- \\
		$R_{a}$ (L/hr) & 720 & 72 \\
		$D_{sa}$ (min) & 12 & 1 \\
		$D_{pa}$ (min) & 4 & 0.3 \\
		$V_{a}$ (km/hr) & 16 & 4 \\
		\bottomrule
	\end{tabular}
\end{table}

\begin{table}
	\caption{Replenishment Point Parameters for Scenario 2}
	\label{t:param_p2}
	\centering
	\begin{tabular}{ c  c  c }
		\toprule
		Parameter & Mean & Standard Deviation \\
		\midrule
		$D_{sr}$ (min) & 12 & 2 \\ 
		$D_{pr}$ (min) & 6 & 1 \\ 
		$R_{r}$ (L/hr) & 12000 & 1200 \\ 
		\bottomrule
	\end{tabular}
\end{table}

\begin{table}
	\caption{User Agent Parameters for Scenario 2}
	\label{t:param_u2}
	\centering
	\begin{tabular}{ c c  c  c }
		\toprule
		Type & Parameter & Mean & Standard Deviation \\
		\midrule
		Small (S) & $c_{u}$ (L) & 480 & -- \\ 
		& $R_{u}$ (L/hr) & 10 & 1 \\ 
		\midrule
		Medium (M) & $c_{u}$ (L) & 600 & -- \\ 
		& $R_{u}$ (L/hr) & 12.5 & 1.25 \\ 
		\midrule
		Large (L)& $c_{u}$ (L) & 720 & -- \\ 
		& $R_{u}$ (L/hr) & 15 & 1.5 \\ 
		\midrule
		Extra Large (XL) & $c_{u}$ (L) & 960 & -- \\ 
		& $R_{u}$ (L/hr) & 20 & 2 \\ 
		\bottomrule
	\end{tabular}
\end{table}

%
%

\begin{table}
	\caption{User Agent Types for Scenario 2}
	\label{t:param_u3}
	\centering
	\begin{tabular}{ c  c  c  c  c  c c c c c c}
		\toprule
		Agent & 1 & 2 & 3 & 4 & 5 & 6 & 7 & 8 & 9 & 10 \\
		Size & S & L & S & M & M & S & M & L & XL & L\\
		\midrule
		Agent & 11 & 12 & 13 & 14 & 15 & 16 & 17 & 18 & 19 & 20 \\
		Size & L & XL & M & L & M & S & L & M & L & M\\
		\bottomrule
	\end{tabular}
\end{table}

\subsection{Evaluation of the prediction framework}\label{s:comp_pred}

The prediction method developed in Section \ref{s:pred} was compared with the analytical prediction method previously developed by the authors in \cite{Palmer2013}, using a Monte Carlo generated cost as a benchmark. They were tested in Scenario 1 with 6 user agents and a schedule length of 8 tasks, and in Scenario 2 with the large replenishment agent and a schedule length of 20 tasks. 10,000 random schedules were generated for each scenario with the initial resource levels of all agents randomly initialised to a value between 0\% and 100\% of capacity for each schedule. The Monte Carlo cost was generated using 1,000 samples as this was found to be a good trade-off between error and calculation time in \cite{Palmer2013}. 

Table \ref{t:results_cost_single_1} shows the error of the proposed and previous methods in comparison to the Monte Carlo method. As can be seen, the proposed method has significantly less error than the previous method, particularly in Scenario 2. A significant source of the error in Scenario 2 for the previous method was from the particular approximation of the inverse Gaussian distributed random variable that was used. As the standard deviation of the velocity of the replenishment agent is very high compared to the mean, the resultant inverse Gaussian distribution for the travel time is heavily skewed. As was shown in Figure \ref{f:inverse}, the old approximation of the inverse Gaussian distribution overestimates the probability in the left tail of the distribution in these cases, resulting in the cost being underestimated. The new approximation does not overestimate this probability to the same extent, and consequently produces significantly better results. 

The comparison accuracy shows how effective each method is at discriminating between schedules. This is the most important aspect of the prediction method---it must be able to accurately discriminate between schedules for it to be effective when used within a schedule optimisation. The proposed method is accurate in 0.3\% more cases in Scenario 1, and 1\% more cases in Scenario 2. In Scenario 2 in particular, this is a significant improvement. 

\begin{table}
	\caption{Proposed cost method minus Monte Carlo cost method}
	\label{t:results_cost_single_1}
	\centering
	\begin{tabular}{c c  c  c  c }
		\toprule
		Scenario & Method  & Mean & Standard  & Comparison \\
		&  & ($\times 10^{-3}$) & Deviation  & Accuracy  \\
		& & & ($\times 10^{-3}$) & \\
		\midrule
		1 & Previous from \cite{Palmer2013} & 3.18 & 3.11 & 99.3\% \\
		1 & Proposed & 0.08 & 1.52 & 99.6\% \\
		\midrule
		2 & Previous from \cite{Palmer2013} & -27.8 & 4.94 & 98.4\% \\
		2 & Proposed & -1.96 & 1.92 & 99.4\% \\
		\bottomrule
	\end{tabular}
\end{table}

The main advantage of the proposed prediction method over the Monte Carlo method is the computation time. The proposed method took just 7ms in Scenario 1 compared to 678ms for the Monte Carlo method, and in Scenario 2 took 9ms compared to 1.506s for the Monte Carlo method. 

\subsection{Evaluation of the optimisation methods}\label{s:opt_pred}

The optimisation methods tested are summarised below, and the acronyms in parentheses will be used to refer to the methods. 

\begin{itemize}
	\item ATC heuristic (ATC)
	\item Simulated annealing using the developed prediction framework (SA)
	\item Branch and bound ignoring uncertainty (DBB)
	\item Branch and bound incorporating uncertainty through the developed prediction framework (SBB)
\end{itemize}

DBB uses the Monte Carlo method from \cite{Palmer2013} with 1 sample, treating all parameters as certain. The SA and SBB methods incorporate uncertainty through the developed prediction framework. 

\subsubsection{Scenario 1} \label{s:scenario1_opt}

\begin{figure}
	\centering
	\includegraphics[width=0.6\textwidth]{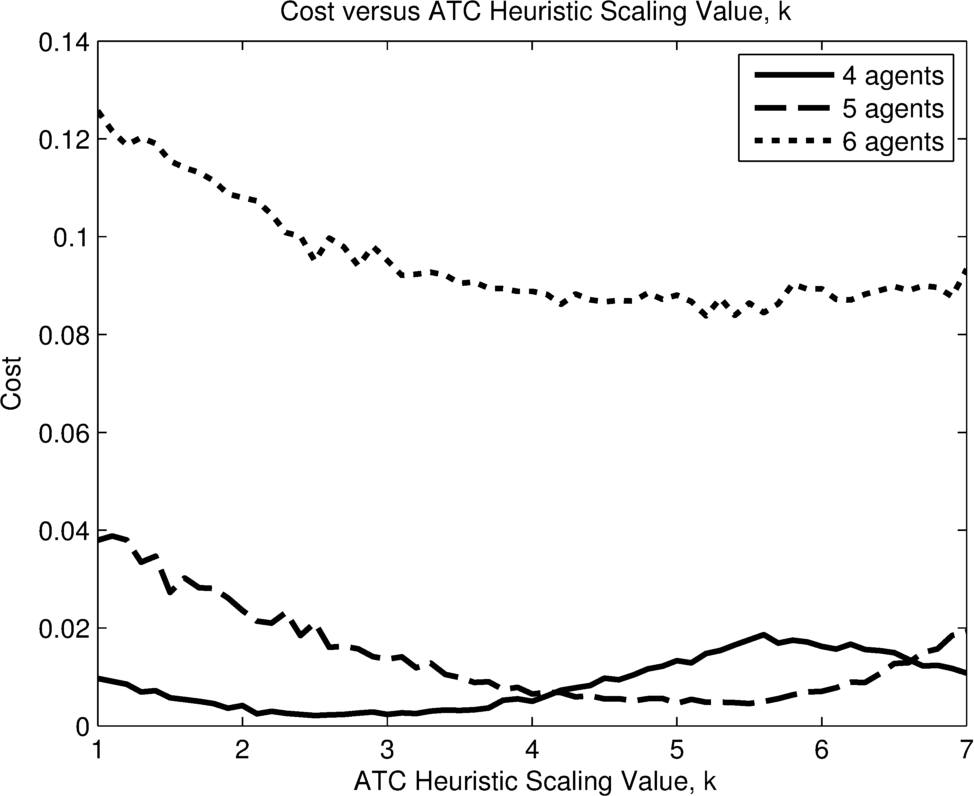}
	\caption{Cost versus $k$ value for the ATC heuristics for the 4-, 5-, and 6-user agent cases in scenario 1. }
	\label{f:ATC}
\end{figure}

Each simulation of Scenario 1 was initialised with random initial resource levels between 50\% and 100\% to simulate realistic in-progress starting conditions, and the simulation lasted for 5 hours of simulated time. The $k$ value for ATC was first tuned by running multiple simulations with values between 1 and 7. As shown in Figure \ref{f:ATC}, the lowest costs were achieved using a $k$ value of approximately 2.5 for the 4-user agent scenario, and approximately 5.5 for the 5- and 6-user agent scenarios. This means that the behaviour is biased more towards the MS rule than the WSPT rule for the 4-user agent scenario in comparison to the 5- and 6-agent scenarios. 

\begin{figure}
	\centering
	\subfloat[4 user agents]{
		\includegraphics[width=0.5\textwidth]{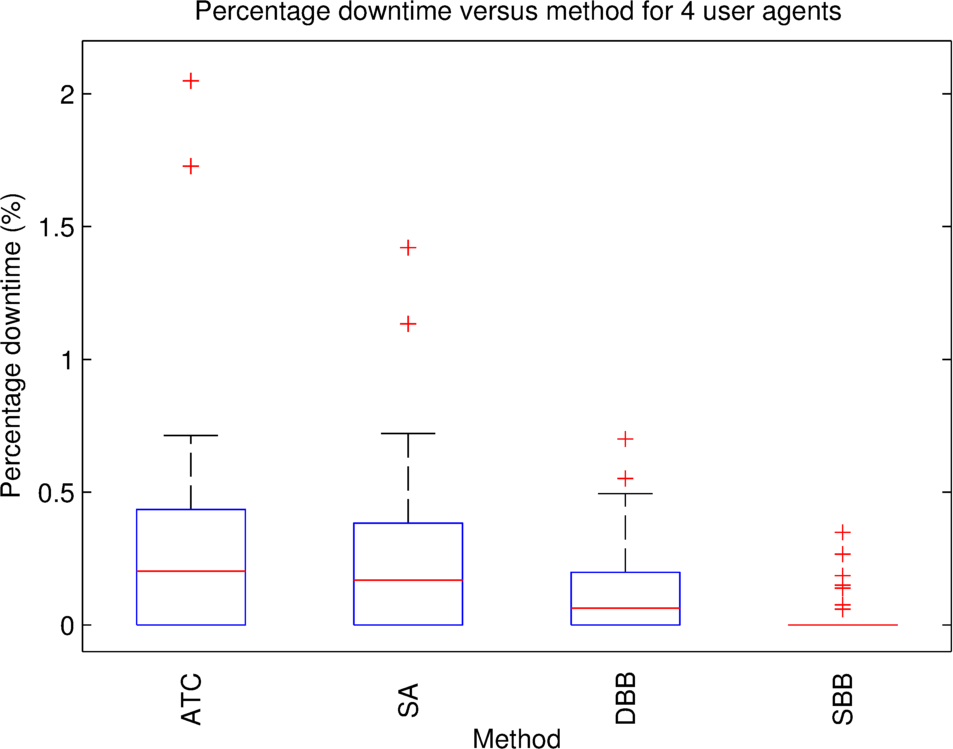}\label{sf:4_agent}
	}
	
	\subfloat[5 user agents]{
		\includegraphics[width=0.5\textwidth]{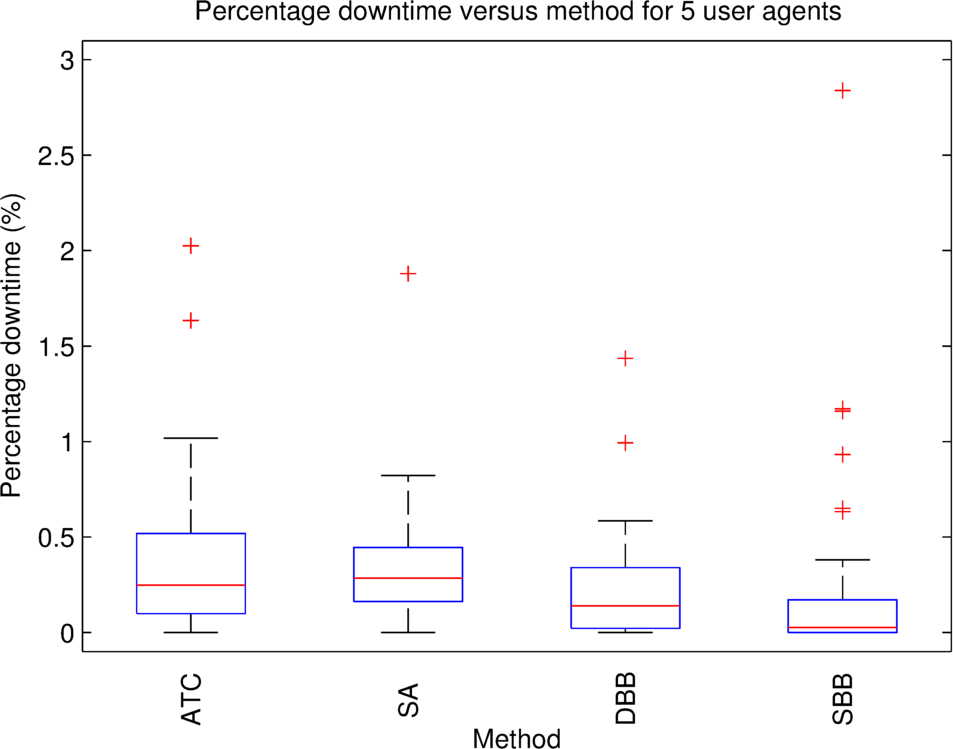}\label{sf:5_agent}
	}
	
	\subfloat[6 user agents]{
		\includegraphics[width=0.5\textwidth]{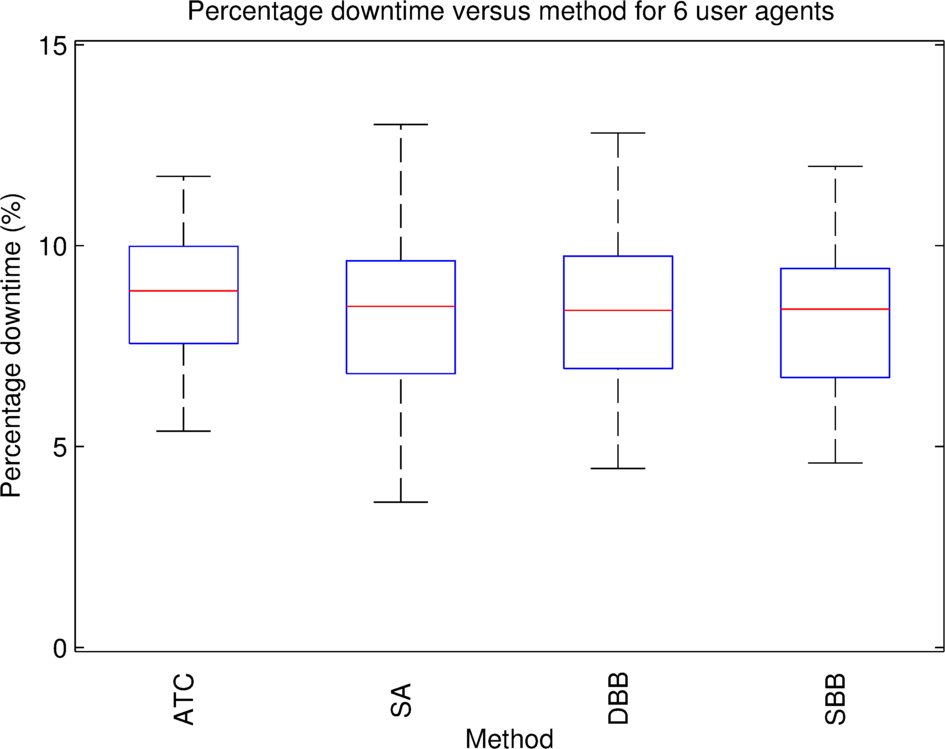}\label{sf:6_agent}
	}
	\caption{Box and whisker plots for the percentage downtime in Scenario 1. Lower results are better. }
	\label{f:scen1}
\end{figure}

\begin{figure}
	\centering
	\subfloat[4 user agents]{
		\includegraphics[width=0.5\textwidth]{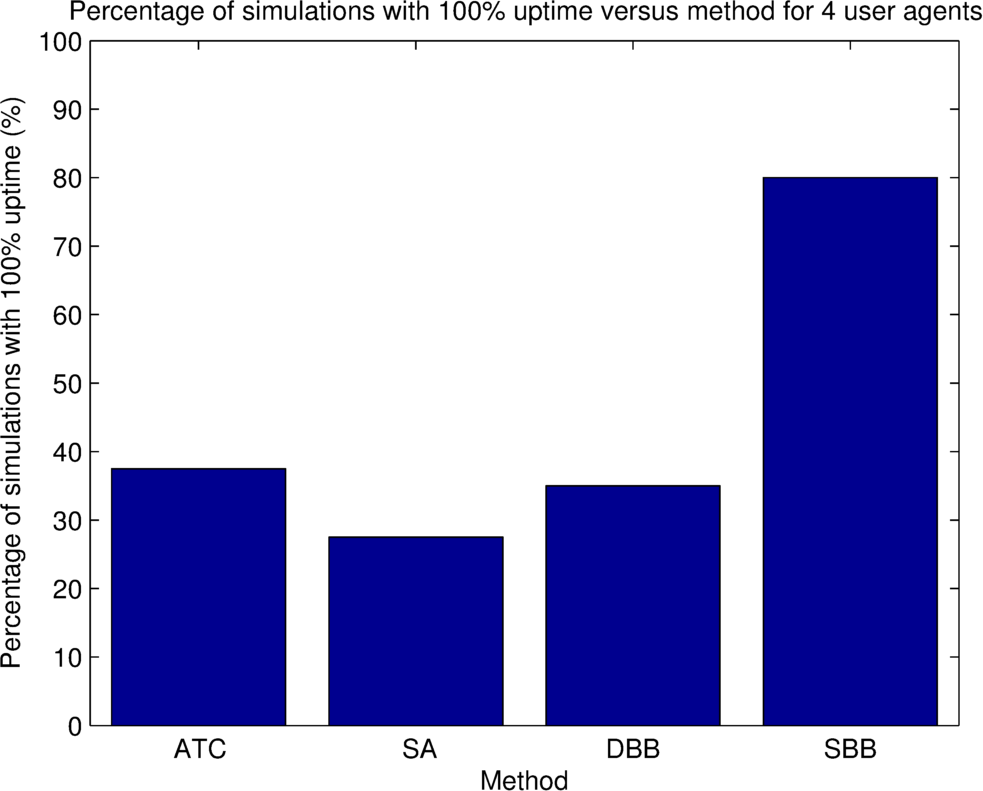}\label{sf:4_agent2}
	}
	
	\subfloat[5 user agents]{
		\includegraphics[width=0.5\textwidth]{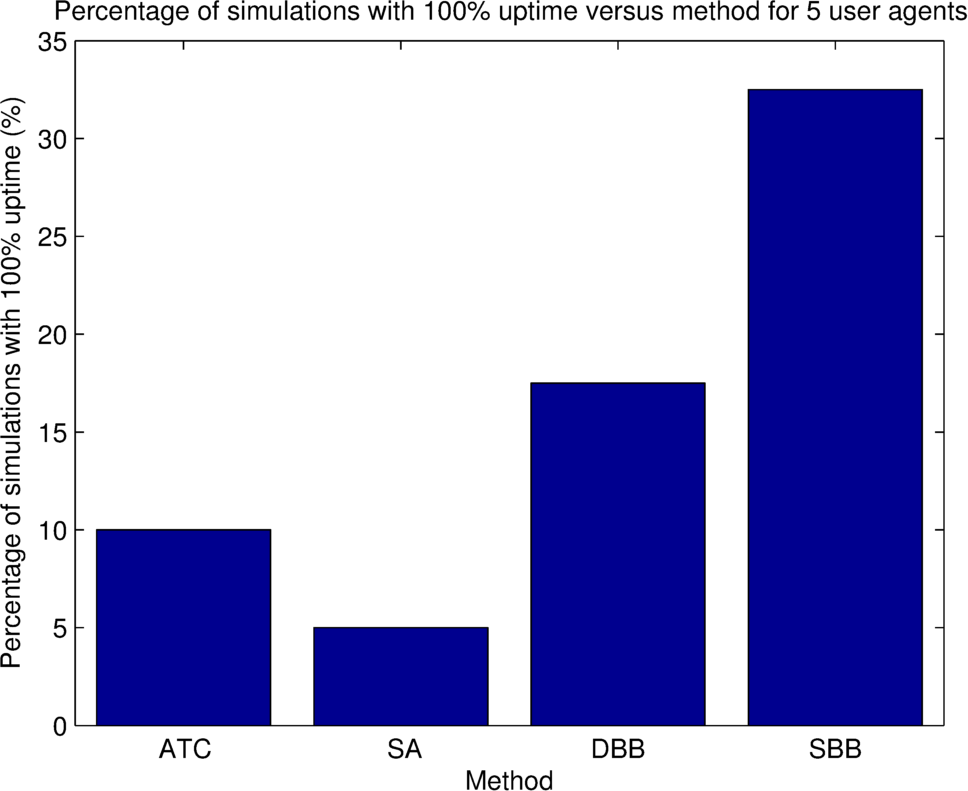}\label{sf:5_agent2}
	}
	\caption{Percentage of simulation runs with 100\% uptime in Scenario 1. Higher results are better. The 6-user agent scenario results are omitted as no method was able to achieve 100\% uptime in any of the simulations. }
	\label{f:scen1_2}
\end{figure}

Each optimisation method was tested 40 times to account for the variability due to the stochastic nature of the simulation. The SA, DBB, and SBB methods were tested using a schedule length of $n+3$ tasks, where $n$ is the number of user agents in the system. Using shorter schedule lengths than this can lead to undesirable behaviour; for further discussion, see \cite{Palmer2014a}. The percentage downtime results from this scenario are shown in Figure \ref{f:scen1}, and Figure \ref{f:scen1_2} shows the percentage of simulations in which none of the user agents exhausted their supply of the resource. As can be seen, the proposed SBB method consistently produced the lowest downtime in the 4- and 5-user agent scenarios. In the 6-user agent scenario, the SBB and DBB methods produced almost identical performance. In over-utilised scenarios like this, the distributions for the downtime of the user agents are predominantly in the positive domain. The expected cost of these distributions is therefore very close to the mean value, and hence very similar to the result returned by the deterministic framework used in DBB. The SA method struggled to find good schedules, providing only a minimal improvement to the initial schedule generated by the ATC heuristic. 

\begin{figure}
	\centering
	\includegraphics[width=0.6\textwidth]{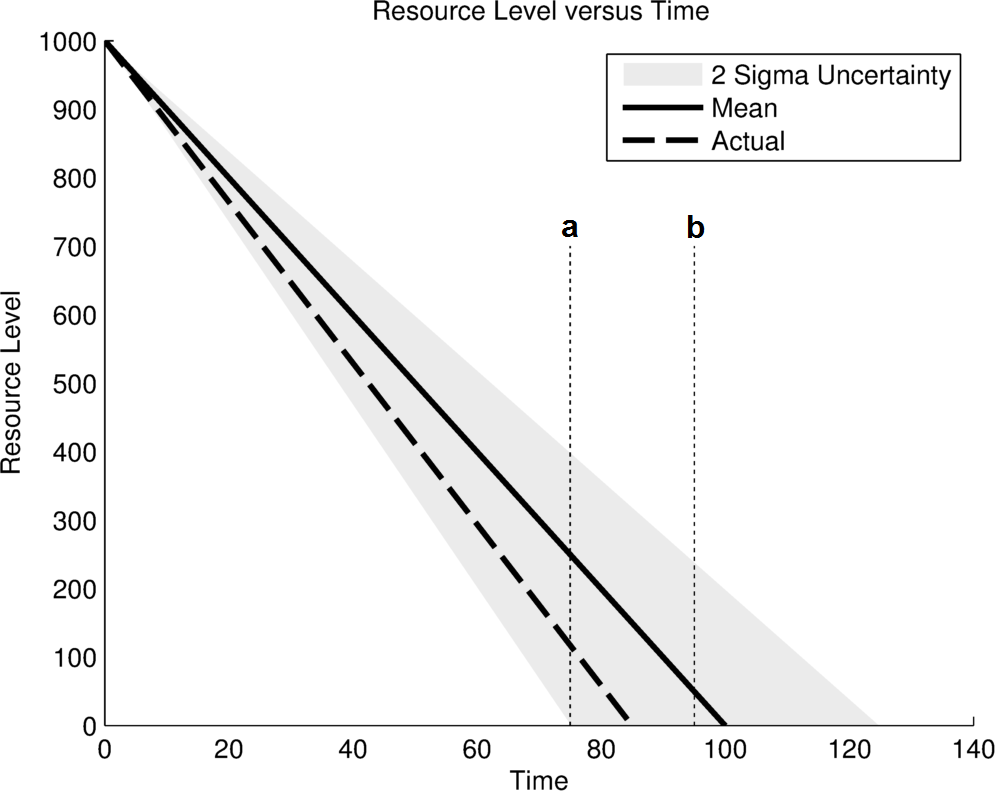}
	\caption{Level of a user agent showing the predicted resource level, an uncertainty of two standard deviations (2 sigma), and the actual resource level. Replenishing at point \textit{a} is selected when considering uncertainty, while the two points cannot be differentiated if uncertainty is not considered. }
	\label{f:uncertainty}
\end{figure}

The main advantage of the SBB method over DBB is highlighted by the percentage of simulations in which none of the user agents incurred downtime. In the 4- and 5-user agent scenarios, the schedules tested by DBB will frequently have a cost of zero. This means that, in many cases, DBB is unable to differentiate between these schedules and consequently relies on the priorities generated by the ATC heuristic to select a good schedule. The proposed prediction framework used by SBB will never result in a cost of zero as there is always some risk associated with each schedule. This is illustrated in Figure \ref{f:uncertainty}. SBB will find that selecting a schedule that will replenish the user agent at point $a$ is significantly less risky than at point $b$, whereas DBB will return a cost of zero for both schedules and is unable to differentiate between them. Therefore, point $b$ may be chosen sometimes by DBB, leading to an incurred cost when the actual resource level is as shown in Figure \ref{f:uncertainty}. 

The calculation times for the various optimisation methods are detailed in Table \ref{t:scen1calctime}. The ATC heuristic is the fastest of the methods, taking a fraction of a second in all cases, while SA consistently takes several seconds. DBB computes a solution very quickly in the 4- and 5-user agent scenarios because it can find a zero-cost schedule---it will usually find a zero-cost schedule in the first few schedules examined and will return this immediately. SBB takes longer in these cases as it searches through the entire tree. In the 6-user agent scenario, DBB is unable to find a zero-cost schedule and spends significantly longer searching through the tree. In this case, SBB is approximately 5 times slower than DBB. This increase reflects the increased computational requirement of the proposed prediction framework over frameworks that ignore uncertainty.

\begin{table}
	\caption{Calculation times in seconds for Scenario 1}
	\label{t:scen1calctime}
	\centering
	\begin{tabular}{c c  c  c  c }
		\toprule
		Number of agents, $n$ & ATC  & SA & DBB  & SBB \\
		\midrule
		4 & 1.37\mbox{\sc{e}-}4 & 6.55 & 2.80\mbox{\sc{e}-}2 & 0.884 \\
		5 & 1.48\mbox{\sc{e}-}4 & 7.96 & 4.60\mbox{\sc{e}-}2 & 2.51 \\
		6 & 1.84\mbox{\sc{e}-}4 & 9.68 & 7.46 & 37.8 \\
		\bottomrule
	\end{tabular}
\end{table}

\subsubsection{Scenario 2}

Each optimisation method was tested 40 times in this scenario, with initial conditions between 50\% and 100\% randomly selected and each simulation lasting for 9 days of simulated time. ATC scaling values, $k$, of 3 for the large and medium sized replenishment agents, and 5 for the small replenishment agent, were found to give good behaviour. Given the large size of this scenario, calculating an optimal schedule is not feasible as there are over $10^{32}$ combinations when a schedule of 25 tasks is considered. This is where the anytime behaviour of branch and bound is a significant benefit over the A* method used in \cite{Palmer2014a}. This anytime nature was exploited in two ways---the optimisation depth of the algorithm was varied, and a hard limit of 10,000 nodes was placed on the size of the solution tree. 

\begin{figure}
	\centering
	\includegraphics[width=0.6\textwidth]{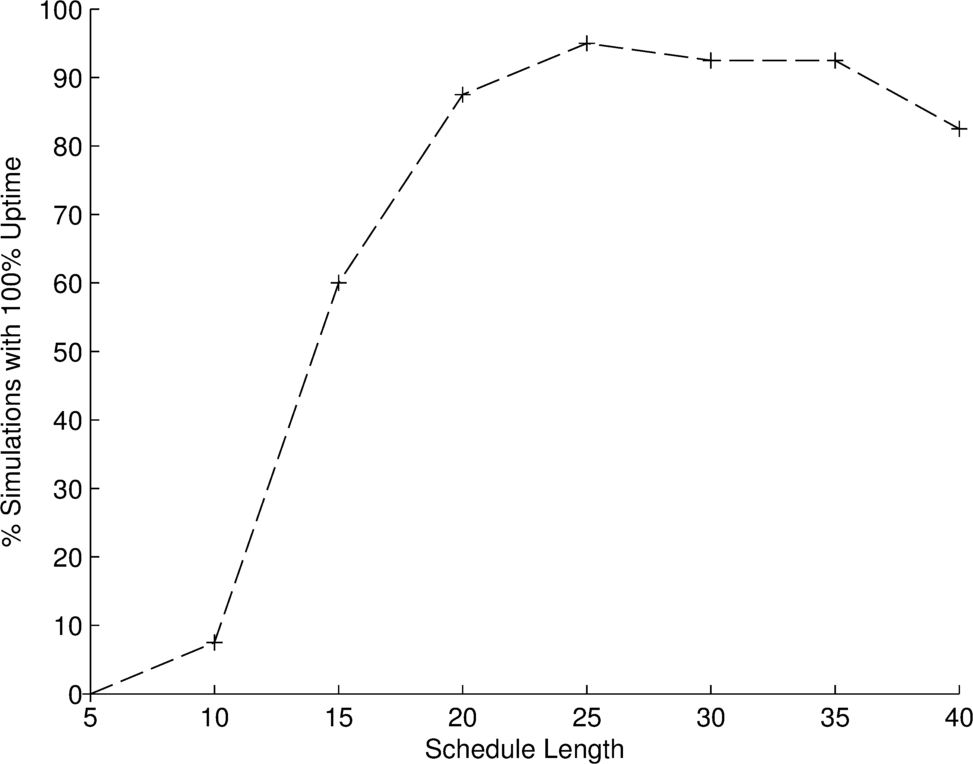}
	\caption{Percentage of simulations with 100\% uptime for various schedule lengths using SBB optimising the first two tasks in scenario 2. Higher results are better. }
	\label{f:schedule_length}
\end{figure}

Selecting the schedule length is an important aspect of this problem. Using a short schedule length can lead to myopic behaviour, while using a long schedule length exponentially increases the size of the solution tree. When using an optimisation depth that is smaller than the schedule length, the remaining tasks in the schedule are selected by the ATC heuristic. If the schedule length is too long in comparison to the optimisation depth, the suboptimal choices of the heuristic can also reduce the benefit of using a longer schedule length. Figure \ref{f:schedule_length} shows the percentage of simulations with 100\% uptime using the SBB algorithm with an optimisation depth of two tasks as a function of the schedule length. As can be seen, the best performance is achieved at a schedule length of 25 tasks. 

\begin{figure}
	\centering
	\subfloat[Large replenishment agent]{
		\includegraphics[width=0.5\textwidth]{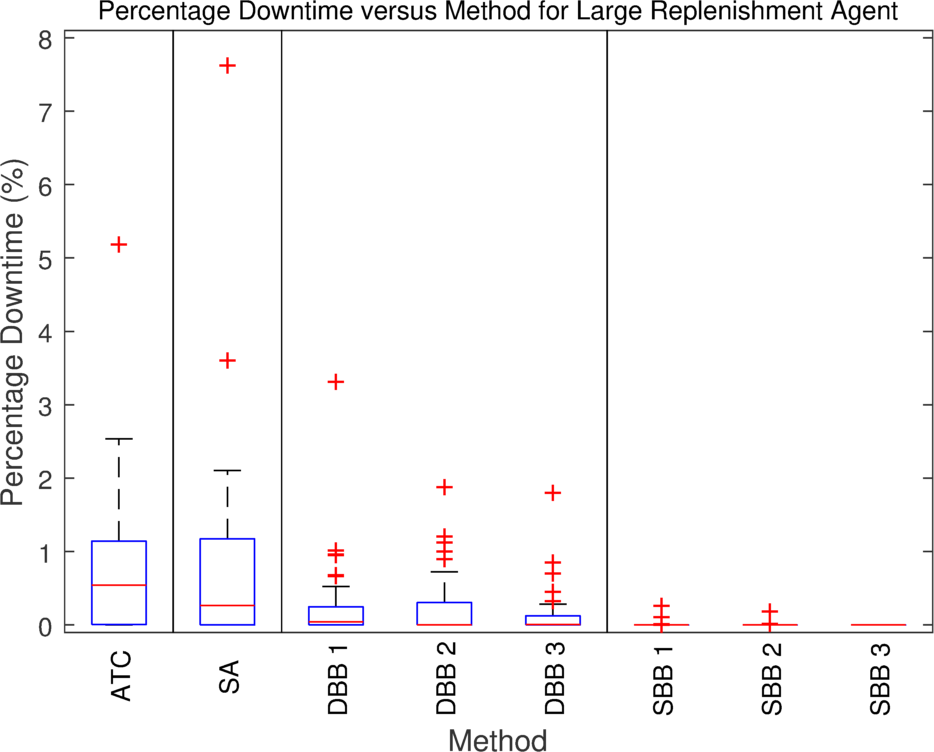}\label{sf:scen2_1}
	}
	
	\subfloat[Medium replenishment agent]{
		\includegraphics[width=0.5\textwidth]{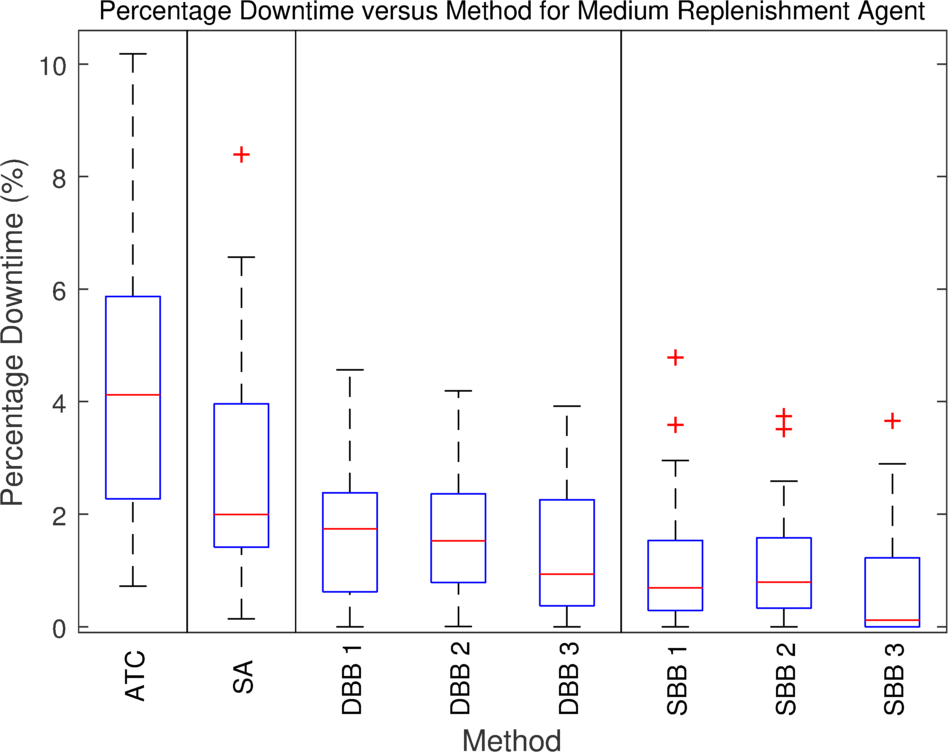}\label{sf:scen2_2}
	}
	
	\subfloat[Small replenishment agent]{
		\includegraphics[width=0.5\textwidth]{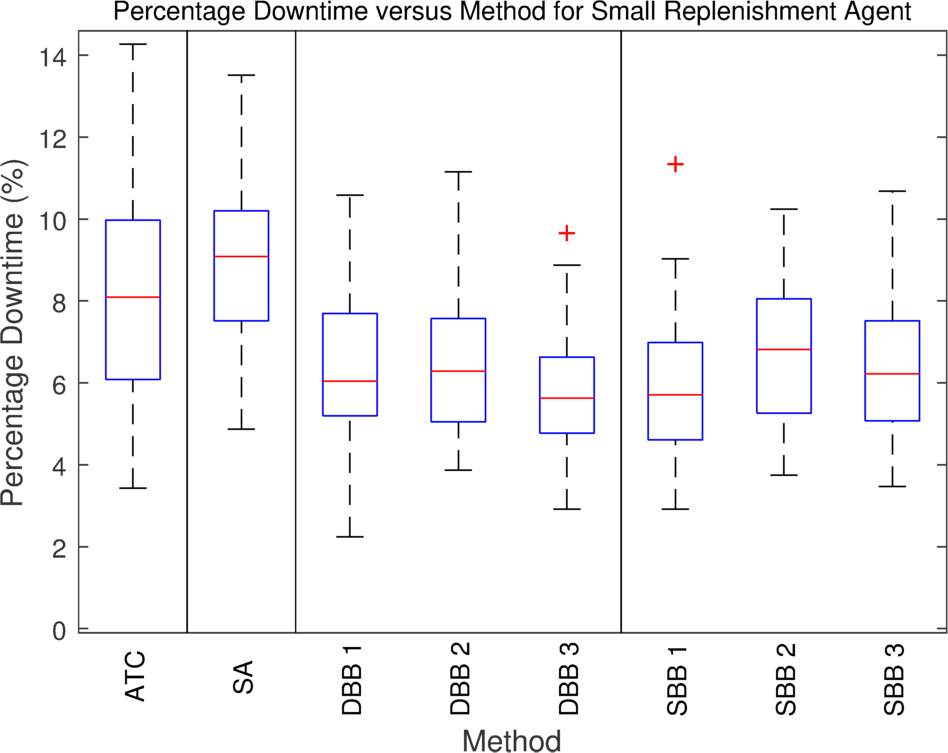}\label{sf:scen2_3}
	}
	\caption{Box and whisker plots for the percentage downtime in Scenario 2. Lower results are better. The number next to the DBB and SBB methods is the optimisation depth.}
	\label{f:scen2results}
\end{figure}

\begin{figure}
	\centering
	\subfloat[Large replenishment agent]{
		\includegraphics[width=0.5\textwidth]{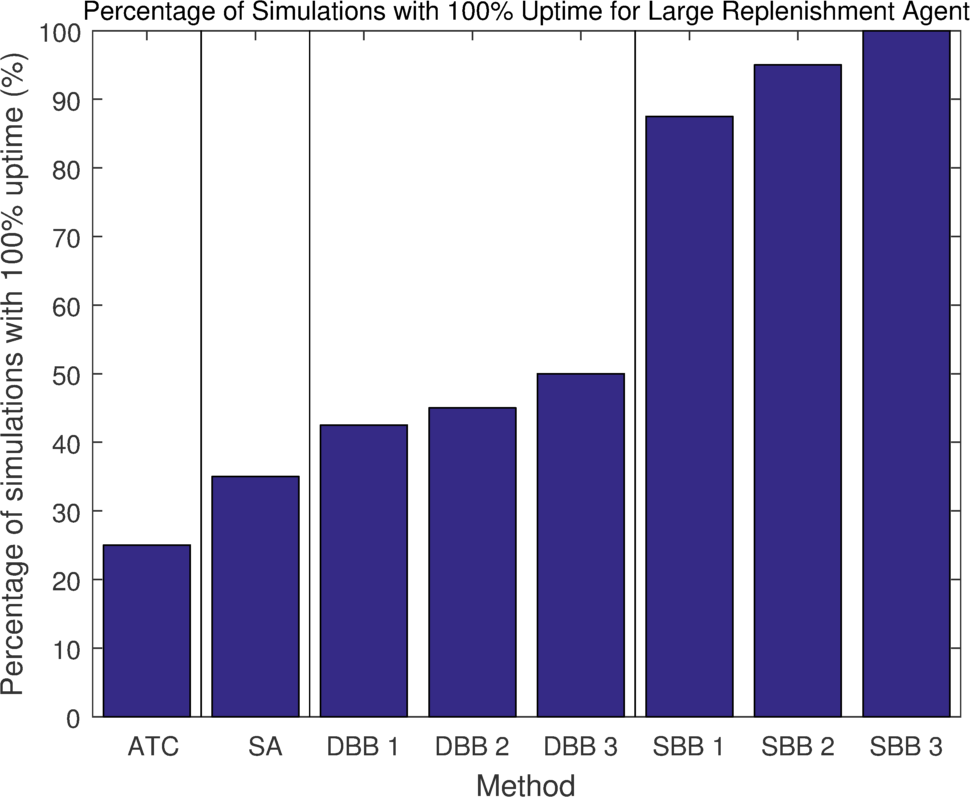}\label{sf:scen2_up_1}
	}
	
	\subfloat[Medium replenishment agent]{
		\includegraphics[width=0.5\textwidth]{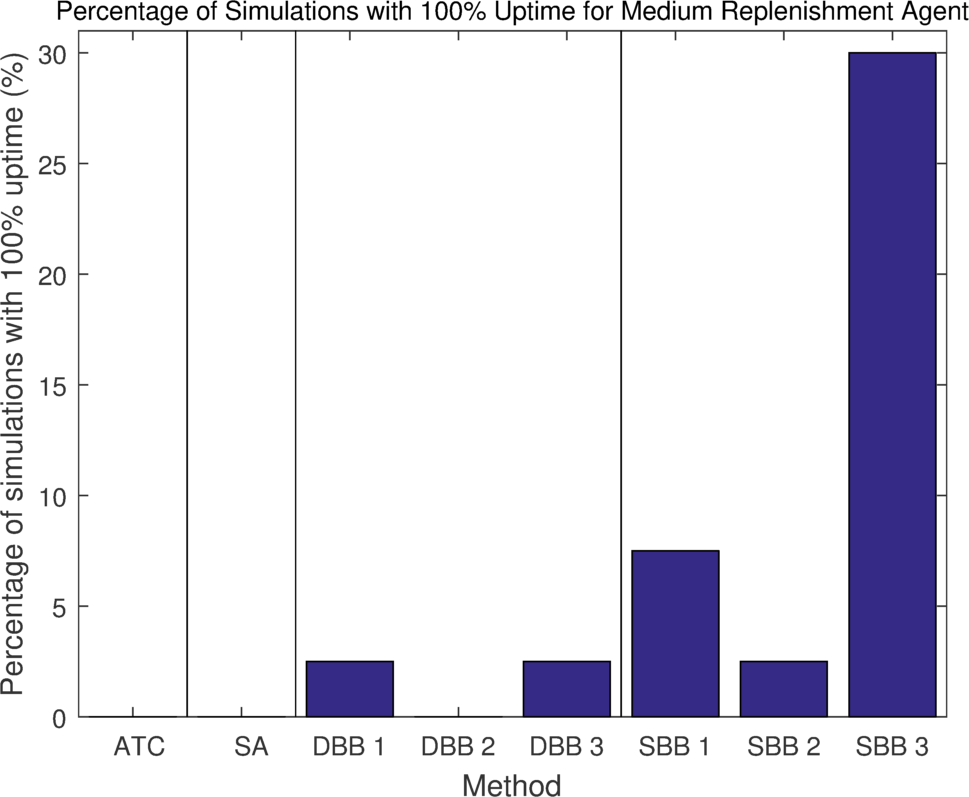}\label{sf:scen2_up_2}
	}
	\caption{Percentage of simulation runs with 100\% uptime in Scenario 2. Higher results are better. The number next to the DBB and SBB methods is optimisation depth. The small replenishment agent scenario results are omitted as no method was able to achieve 100\% uptime in any of the simulations. }
	\label{f:scen2results2}
\end{figure}

Figures \ref{f:scen2results} and \ref{f:scen2results2} show the results for the optimisation methods in the second scenario. SA, DBB, and SBB were tested using a schedule length of 25 tasks, with optimisation depths of 1, 2, and 3 used by DBB and SBB. These results broadly mirror those in Scenario 1, with the proposed SBB method producing the best results. The benefit of the directed optimisation of the branch and bound methods on an initial schedule generated by the ATC heuristic is evident here. Even if only the first task is optimised, DBB and SBB provide a huge benefit over the ATC heuristic. In both the large and medium replenishment agent cases, SBB clearly outperformed DBB. This is highlighted by the results in Figure \ref{f:scen2results2}. In the small replenishment agent case, SBB and DBB produced similar results, corroborating the findings from Scenario 1. As these methods are used within an MPC-like framework, it is beneficial to focus the optimisation on earlier tasks within a schedule. SA struggled to find good schedules in this scenario as it spread the optimisation efforts across the entire schedule. As a result, it was unable to sufficiently explore the search space to yield much improvement over the ATC heuristic. 

Larger scenarios than this can be handled by the proposed approach, but may require additional planning time to achieve the significant improvements seen in the two scenarios examined in this paper. This is because the number of possible schedules increases exponentially as the number of user agents is increased. If the search time is held constant, then the solution qualities from the branch and bound methods will gradually decrease with the number of user agents. In the limit, the behaviour of the algorithm will become dominated by the heuristic used to select tasks for generating complete schedules out to the planning horizon. 

The calculation times for each method are shown in Table \ref{t:scen2calctime}. These results follow similar trends to the results in Scenario~1, with the ATC heuristic computing very quickly, and the branch and bound methods taking the longest. In the large and medium replenishment agent cases, DBB has a short calculation time compared to SBB as it quickly finds zero-cost schedules. In the small replenishment agent case, the calculation times of DBB and SBB are much closer together. Many opportunities exist for improving the calculation times of these methods including using a language such as C, using a parallel implementation of branch and bound, and storing more data in the tree structure to reduce the computation required at each node. Using conservative estimates, speed increases of at least 100 times are feasible, giving potential sub-second calculation times for the branch and bound methods.

\begin{table}
	\caption{Calculation times in seconds for Scenario 2}
	\label{t:scen2calctime}
	\centering
	\begin{tabular}{c c  c  c  c c}
		\toprule
		Agent size & ATC  & SA & DBB 1 & DBB 2 & DBB 3 \\
		\midrule
		Large & 1.61\mbox{\sc{e}-}3 & 13.5 & 0.664 & 0.674 & 0.668 \\
		Medium & 1.50\mbox{\sc{e}-}3 & 13.9 & 0.671 & 0.639 & 0.638 \\
		Small & 1.64\mbox{\sc{e}-}3 & 13.1 & 2.38 & 24.2 & 29.8 \\
		\midrule
		Agent size & SBB 1 & SBB 2  & SBB 3 & &\\
		\midrule
		Large & 1.38 & 11.9 & 12.2 &  & \\
		Medium & 1.88 & 21.7 & 39.5 &  & \\
		Small & 1.68 & 20.1 & 48.6 &  & \\
		\bottomrule
	\end{tabular}
\end{table}

\section{Conclusion} \label{s:conc}

Research on replenishment and collection scenarios has thus far not taken into account long term considerations, such as the limited capacity of the replenishment agent and replenishing user agents multiple times, that are critical for achieving persistent autonomy. More importantly, the examined literature has treated these scenarios as deterministic, ignoring the uncertainty that is inherent in realistic scenarios. These aspects play an important role in the optimisation process, and the SCAR scenario was developed to specifically address these shortcomings. 

This paper proposed a novel framework for incorporating uncertainty when predicting the outcome of the schedule for the replenishment agent in a SCAR scenario, and developed a branch and bound method that used the prediction framework to optimise the schedule of the replenishment agent. Improved Gaussian approximations enabled the proposed prediction framework to outperform an existing framework. The branch and bound approach using this framework was then shown to outperform the ATC heuristic, a simulated annealing meta-heuristic, and a branch and bound approach ignoring uncertainty, in both a small and large scenario. In the large scenario, the anytime characteristic of branch and bound was exploited to find good schedules within a reasonable length of time by varying the optimisation depth of the algorithm. Tasks beyond the optimisation depth were selected using the ATC heuristic, enabling the use of a long schedule length to reduce myopic decision making. 

An interesting avenue of future work is considering systems with multiple replenishment agents or multiple resources, as the size of the search space will be substantially larger than for the single replenishment agent or single resource scenarios, thus requiring more efficient optimisation methods in order to select appropriate tasks. One possible method could be to cluster the user agents so that the problem reduces to multiple single-replenishment agent optimisations. Other areas of future work include considering uncertainty on the current state of the system, assessing the robustness of the methods to changes in the underlying probability distributions, and developing methods for dynamically adapting the $k$ values for the ATC heuristic. 

\section*{Acknowledgements}

This work was supported by the Rio Tinto Centre for Mine Automation and the Australian Centre for Field Robotics, University of Sydney, Australia.



\bibliographystyle{elsarticle-num} 
\bibliography{ref}


%
%
%
\end{document}